%% file: main.tex
\documentclass[table]{gtech}
\PassOptionsToPackage{table, usenames, dvipsnames}{xcolor}
\usepackage{xcolor}
\usepackage{amssymb}
\usepackage{multirow}
\usepackage{bigdelim}
\usepackage{longtable}
\usepackage{tabularray}
\usepackage{wrapfig}
\usepackage[most]{tcolorbox}
\usepackage{url}
\usepackage{float}
\usepackage{datatool}
\usepackage{enumitem}
\usepackage{subcaption} %
\usepackage{caption}

\RequirePackage{tgpagella} %
\RequirePackage{mathpazo}  %
\RequirePackage{inconsolata} %
\usepackage{makecell}

\usepackage{adjustbox}
\usepackage{tablefootnote}
\usepackage{threeparttable}

\usepackage{booktabs} %
\usepackage{array}    %
\newcolumntype{C}[1]{>{\centering\arraybackslash}m{#1}}

\usepackage[utf8]{inputenc} %
\usepackage[T1]{fontenc}    %
\usepackage{hyperref}       %
\usepackage{url}            %
\usepackage{booktabs}       %
\usepackage{amsfonts}       %
\usepackage{nicefrac}       %
\usepackage{microtype}      %
\usepackage{xcolor}         %
\usepackage{xspace}
\usepackage{amsthm}   %
\usepackage{amsmath}  %
\usepackage{amssymb}  %
\usepackage{bm}       %

\theoremstyle{definition}

\renewcommand{\title}[1]{\newcommand{\titlelist}{{\huge\fontfamily{optimistic}\selectfont #1}}}

\newcommand{\ignore}[1]{}

\usepackage{arydshln}
\definecolor{CQColor}{rgb}{0.0,0.0,1.0} %

\usepackage{makecell}
\usepackage{siunitx}

\usepackage{graphicx}
\usepackage{colortbl}
\usepackage{amssymb}
\usepackage{pifont}
\usepackage{booktabs,multirow}
\usepackage{makecell}
\usepackage{tabulary}
\usepackage{fontawesome5}
\usepackage{bbding}
\usepackage{multicol}

\newlength\savewidth

\newcommand{\cmark}{\ding{51}}
\newcommand{\xmark}{\ding{55}}

\usepackage{biolinum}
\newcommand{\NAME}{{\fontfamily{LinuxBiolinumT-LF}\selectfont\textbf{Falcon-X}}\xspace}
\newcommand{\icon}{\raisebox{-2pt}{\includegraphics[width=1.0em]{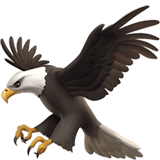}}\xspace}

\title{
\huge{\icon \textcolor[HTML]{0369ff}{\NAME:} A Time Series Foundation Model\\for Heterogeneous Multivariate Modeling}
}

\author[*]{Yiding Liu}
\author[*]{Yifan Hu}
\author[*]{Hongjie Xia}
\author[*]{Peiyuan Liu}
\author[]{Hongzhou Chen}
\author[]{Xilin Dai}
\author[\dagger]{Zewei Dong}
\author[]{Jiangming Yang}

\contribution[*]{Equal Contribution}
\contribution[\dagger]{Corresponding Author,\ \  
\faEnvelope[regular] \{yiding.lyd, zewei.dong\}@ant-intl.com}

\abstract{
  Time series foundation models (TSFMs) are transforming the forecasting paradigm through large-scale cross-domain pretraining. However, most existing TSFMs remain univariate, and recent efforts to enable cross-variate modeling still operate directly within the raw variate space. This design introduces fundamental limitations in semantic alignment and relational expressivity. 
  Specifically, raw-space group mixing lacks a dedicated mechanism to align heterogeneous physical quantities, while standard non-negative attention fails to capture the complex synergistic and antagonistic interactions ubiquitous in real-world systems.
  To address these challenges, we propose \NAME, decouples variates from the raw space and maps them into a unified latent prototype space.
  \NAME employs a Unified Prototype Diff-Attention mechanism that explicitly evaluates both positive and negative semantic affinities to explicitly align heterogeneous variates. Cross-variate interactions are then efficiently performed within this shared space via Latent Entity Attention, naturally facilitating zero-shot structural transfer. Finally, a Variate Reassembly Router robustly reconstructs variate-specific trajectories via a request-and-dispatch mechanism. Extensive evaluations on the GIFT-Eval and fev-bench benchmarks demonstrate that \NAME achieves excellent forecasting performance, offering a principled and scalable paradigm for complex multivariate environments.
  \NAME is publicly released to support future research.
}

\date{\today\vspace{-1mm}}
\gtechdata[Code]{\url{https://github.com/ant-intl/Falcon-TST}\vspace{-1mm}}
\gtechdata[Model]{\url{https://pypi.org/project/falcon-tst/}}

\begin{document}
\maketitle

\input{sections/1-intro}
\input{sections/2-related_work}
\input{sections/3-method}
\input{sections/4-training_details}
\input{sections/5-experiment}
\input{sections/6-conclusion}

\newpage
\bibliographystyle{assets/plainnat}
\bibliography{main}

\newpage
\appendix
\input{sections/appendix}

\end{document}

%% file: sections/1-intro.tex
\section{Introduction}

Time series forecasting is a fundamental task for understanding dynamic systems and supporting future-oriented decision making. Conventional deep forecasting models are typically trained at the dataset level~\citep{survey}, making them difficult to reuse across domains, sampling frequencies, and variate structures that vary in the real world. Time series foundation models (TSFM) are reshaping this paradigm by pretraining on large-scale cross-domain data and transferring directly to new forecasting tasks~\citep{foundation_survey}, thereby substantially reducing the cost of repeated training and tuning. However, most existing models still take univariate series as the basic modeling unit, extrapolating the future solely from each series' own history. This formulation disconnects the co-evolving relationships that are ubiquitous in real systems, limiting the ability of TSFMs to capture complex multivariate dynamics.

To enable cross-variate modeling, recent TSFMs explore how foundation models can accommodate varying numbers of variates.
As shown in Table\ref{tab:tsfm_comparison}, Moirai-1.0~\citep{woo2024unified} flattens multivariate series into a single sequence, enabling joint attention across time and variates. While straightforward, this design scales poorly with the number of variates, as attention cost increases rapidly in high-dimensional settings.
A more effective alternative is group attention~\citep{cohen2025toto, ansari2025chronos2}, which organizes variates into dataset-, entity-, or task-level groups and confines attention to variates within the same group. This avoids indiscriminate mixing across unrelated samples while allowing a shared backbone to process multivariate inputs with varying dimensionalities. By replacing global all-to-all interaction with structured within-group communication, group attention marks an important step toward scalable multivariate TSFMs.

\begin{figure}[!tb]
    \centering
    \includegraphics[width=\textwidth]{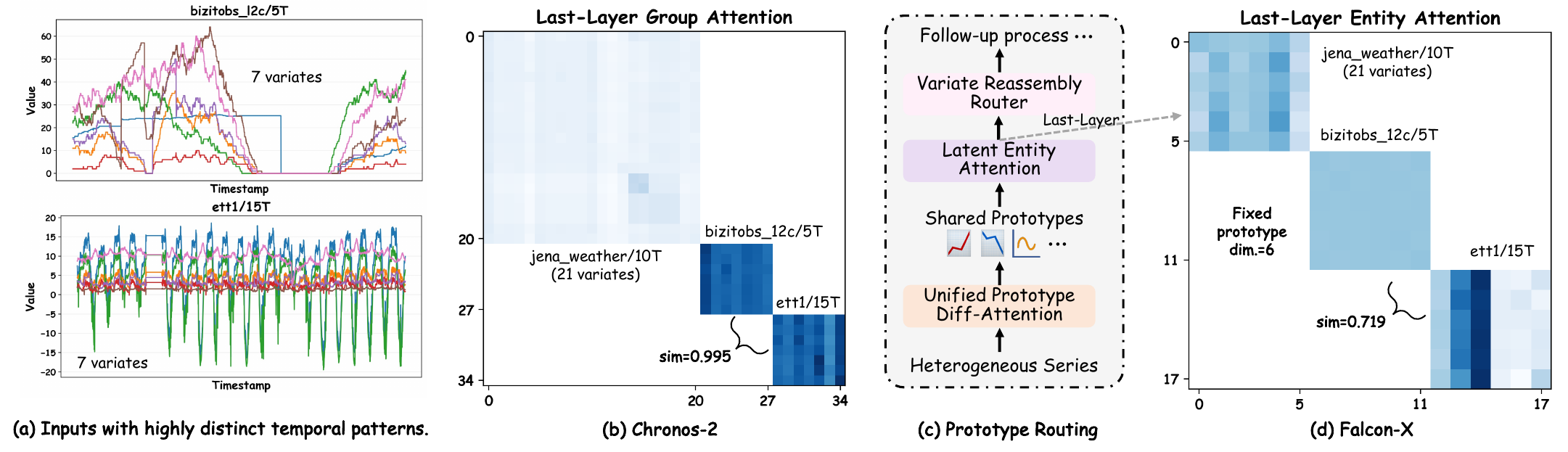}
    \caption{Comparison of multivariate modeling paradigms. (a) Heterogeneous inputs with highly distinct temporal patterns. (b) Group attention produces almost identical attention maps for completely dissimilar inputs, exposing the severe semantic collapse and over-smoothing in the raw variate space. (c-d) In contrast, \NAME projects variates into a latent prototype space, yielding highly discriminative attention maps that accurately capture the underlying dynamics.}
    \label{fig:intro}
\end{figure}

However, group attention still exhibits fundamental limitations in \textbf{\textit{semantic alignment}} and \textbf{\textit{relational expressivity}}.
\ding{182} It operates directly in the raw variate space, controlling which variates can interact but not how their heterogeneous semantics are aligned. In high-dimensional systems, only a small subset of variates typically exhibits strong dependencies, while many others are weakly related or noisy. As shown in Figure \ref{fig:intro}, Dense attention over raw variates can therefore dilute meaningful signals and promote dataset-specific correlations as if they were transferable patterns. 
Moreover, different datasets share the same Transformer backbone, yet their variates often correspond to entirely different physical quantities and dynamics. Without an dedicated alignment space, the model must absorb such heterogeneity implicitly in its parameters, making cross-domain transfer a by-product of parameter sharing rather than an principled process of organizing reusable temporal structures.
\ding{183} Existing attention mechanisms have limited relational expressivity. In real-world systems, cross-variate dependencies often involve both synergistic and antagonistic interactions. However, current attention formulations  primarily capture aggregative effects, lacking the ability to represent opposing dynamics and more complex interaction patterns.

In this paper, we address these limitations by decoupling physical variates from the latent space used for cross-variate interaction. Instead of mixing raw variates, we map them into a shared fixed-dimensional latent prototype space, 
where interactions are mediated by pairs of learnable prototypes that explicitly capture both positive and negative semantic affinities.
These prototypes absorb recurring temporal structures across datasets and filter out variate-specific noise, enabling the model to learn reusable cross-domain patterns in an unified aligned space. 
This alignment establishes a common semantic coordinate system for heterogeneous variates, making cross-domain interactions more structured, transferable, and scalable. It also replaces dense variate-level attention with lightweight variate-to-prototype interaction, substantially improving efficiency while preserving cross-variate modeling capacity.

% This alignment provides a common coordinate system for heterogeneous variates, making knowledge sharing more structured and transferable. It also replaces dense variate-level attention with lightweight variate-to-prototype interaction, improving scalability while preserving cross-variate modeling capacity.

Technically, we propose \NAME, a novel encoder-only TSFM with 591 million parameters materializing this latent prototype paradigm for heterogeneous multivariate forecasting.
At its core, the Unified Prototype Diff-Attention decouples heterogeneous variates into a fixed semantic space, utilizing a differential mechanism to explicitly capture both synergistic and antagonistic variate relationships.
Once aligned, the Latent Entity Attention performs global cross-variate interactions entirely within this unified space, naturally facilitating seamless cross-domain transfer without coupling to raw variate dimensions. 
To project back to the original physical space, a dynamic Variate Reassembly Router robustly reconstructs variate-specific trajectories via a \textit{request-and-dispatch} mechanism and gated residual connections.
Furthermore, \NAME integrates essential instance-wise normalization, patching, and a probabilistic forecasting head to ensure end-to-end stability. 
Extensive experiments on the GIFT-Eval~\citep{aksu2024gift} and fev-bench~\citep{shchur2025fev} validate that \NAME advances the effectiveness in scalable and zero-shot multivariate forecasting.

In a nutshell, our main contribution can be summarized as: 
\begin{itemize}
    \item \textit{\textbf{A Novel Heterogeneous Modeling Paradigm.}} We present a paradigm shift for multivariate time series foundation models, moving from raw-space mixing to a unified latent prototype space. This approach elegantly resolves semantic discrepancies across different datasets, creating a universal coordinate system that naturally facilitates zero-shot knowledge transfer.
    \item \textit{\textbf{Architectural Innovation.}} We propose \NAME, a tailored encoder-only foundation model. It features a Differential Prototype Attention which comprehensively captures both synergistic and antagonistic system dynamics, together with a gated Variate Reassembly Router that adaptively regulates global context fusion for cross-dataset robustness.
    \item \textit{\textbf{Empirical Excellence.}} Through extensive evaluations on comprehensive widely used benchmarks (GIFT-Eval and fev-bench), \NAME consistently achieves superior forecasting performance. The results validate its superior structural adaptability and broad generalization capabilities in complex multivariate environments.
\end{itemize}

%% file: sections/2-related_work.tex
\section{Related Works}

In recent years, the emergence of pre-training techniques has shifted the paradigm of time series forecasting from domain-specific models to the era of foundation models. Early explorations in this domain, such as Time-LLM~\citep{jintime}, primarily focused on tuning third-party large language models to adapt time series tasks. Subsequently, with the accumulation of massive time series data, the trend has transitioned toward learning from scratch, where models are pre-trained directly on large-scale time series data to capture inherent temporal dynamics. Chronos~\citep{ansari2024chronos} frames time series forecasting as a language modeling task by tokenizing real-valued observations into a discrete vocabulary. MOMENT~\citep{goswamimoment} introduces a family of foundation models using a masked prediction task, which allows for generalization across time series tasks through fine-tuning. Inspired by the success of large language models, the generative paradigm built upon decoder-only architectures has become a prominent choice for many mainstream works. These methods typically partition time series into non-overlapping patches, treating each patch as a single token~\citep{nietime}, and leverage decoder-only structures to generate one patch at each step~\citep{liu2024timer, das2024decoder, liu2025moirai, liu2025sundial, auertirex}. 
Despite achieving competitive zero-shot performance, these models primarily rely on the channel independence strategy, which limits them to univariate forecasting and ignores the rich context of multivariate dependencies found in real-world time series data.

Despite these advances, a key challenge of multivariate forecasting remains the unified modeling of heterogeneous time series. Moirai 1.0~\citep{woo2024unified} flattens variates into a single sequence to capture joint interactions. Toto~\citep{cohen2025toto} introduces proportional factorized space-time attention to efficiently model cross-variate dependency. Furthermore, Chronos-2~\citep{ansari2025chronos2} further adopts group attention, facilitating in-context learning by sharing information across related series within flexible groups.
However, these approaches still operate directly within the raw variate space, fundamentally limiting their semantic alignment and relational expressivity. 
Specifically, they primarily capture aggregative effects and struggle to represent the antagonistic dynamics ubiquitous in real-world physical systems. 
To bridge this gap, our \NAME introduces a differential mechanism within an explicitly aligned latent prototype space, enabling the foundation model to systematically capture complex dual dynamics and facilitate robust cross-domain transfer.

\begin{table*}[!t]
\centering
\caption{Comparison of capabilities of TSFMs.}
\label{tab:tsfm_comparison}
\renewcommand{\arraystretch}{2}
\setlength{\tabcolsep}{3.2pt}
\footnotesize
\resizebox{\textwidth}{!}{
\begin{tabular}{@{}cccccccccc@{}}
\toprule
\multirow{2}{*}{\scalebox{1.2}{\textbf{TSFMs}}} 
& \textbf{\NAME} 
& \makecell{\textbf{Moirai 2.0}} 
& \makecell{\textbf{TimesFM-2.5}} 
& \makecell{\textbf{Chronos-2}} 
& \makecell{\textbf{Timer-S1}} 
& \makecell{\textbf{Toto}} 
& \makecell{\textbf{Sundial}} 
& \makecell{\textbf{Moirai 1.0}} 
& \makecell{\textbf{TabPFN-TS}} \\

& \makecell{\textbf{(Ours)}} 
& \citeyearpar{liu2025moirai} 
& \citeyearpar{das2024decoder} 
& \citeyearpar{ansari2025chronos2} 
& \citeyearpar{liu2026timers1} 
& \citeyearpar{cohen2025toto} 
& \citeyearpar{liu2025sundial} 
& \citeyearpar{woo2024unified} 
& \citeyearpar{hoo2025tables} \\
\midrule

\makecell[c]{\textbf{Univariate}} 
& \cmark & \cmark & \cmark & \cmark & \cmark & \cmark & \cmark & \cmark & \cmark \\

\makecell[c]{\textbf{Multivariate}} 
& \cmark & \xmark & \xmark & \cmark & \xmark & \cmark & \xmark & \cmark & \xmark \\

\makecell[c]{\textbf{Cross Learning$^{1}$}} 
& \cmark & \xmark & \xmark & \cmark & \xmark & \xmark & \xmark & \xmark & \xmark \\

\makecell[c]{\textbf{Signed Dependence$^{2}$}} 
& \cmark & \xmark & \xmark & \xmark & \xmark & \xmark & \xmark & \xmark & \xmark \\

\makecell[c]{\textbf{Heterogeneous}\\\textbf{Unification}$^{3}$} 
& \makecell{Prototype\\Routing}  
& \xmark 
& \xmark 
& \makecell{Group\\Mixing}  
& \xmark 
& Fixed 
& \xmark 
& Concat. 
& \xmark \\

\bottomrule
\end{tabular}
}
\vspace{5pt}
\begin{tablenotes}[flushleft]
\footnotesize
\item[] 
\begin{minipage}{\textwidth}
$^{1}$Transferring of universal cross-variate interactive patterns across distinct datasets. \\
$^{2}$Computing both positive and negative affinities to capture synergistic and antagonistic dynamics. \\
$^{3}$Projecting physical variates with varying dimensionalities into a dimension-agnostic latent space.
\end{minipage}
\end{tablenotes}
\end{table*}

%% file: sections/3-method.tex
\section{\NAME}
\label{sec:method}

\begin{figure}[!tb]
    \centering
    \includegraphics[width=\textwidth]{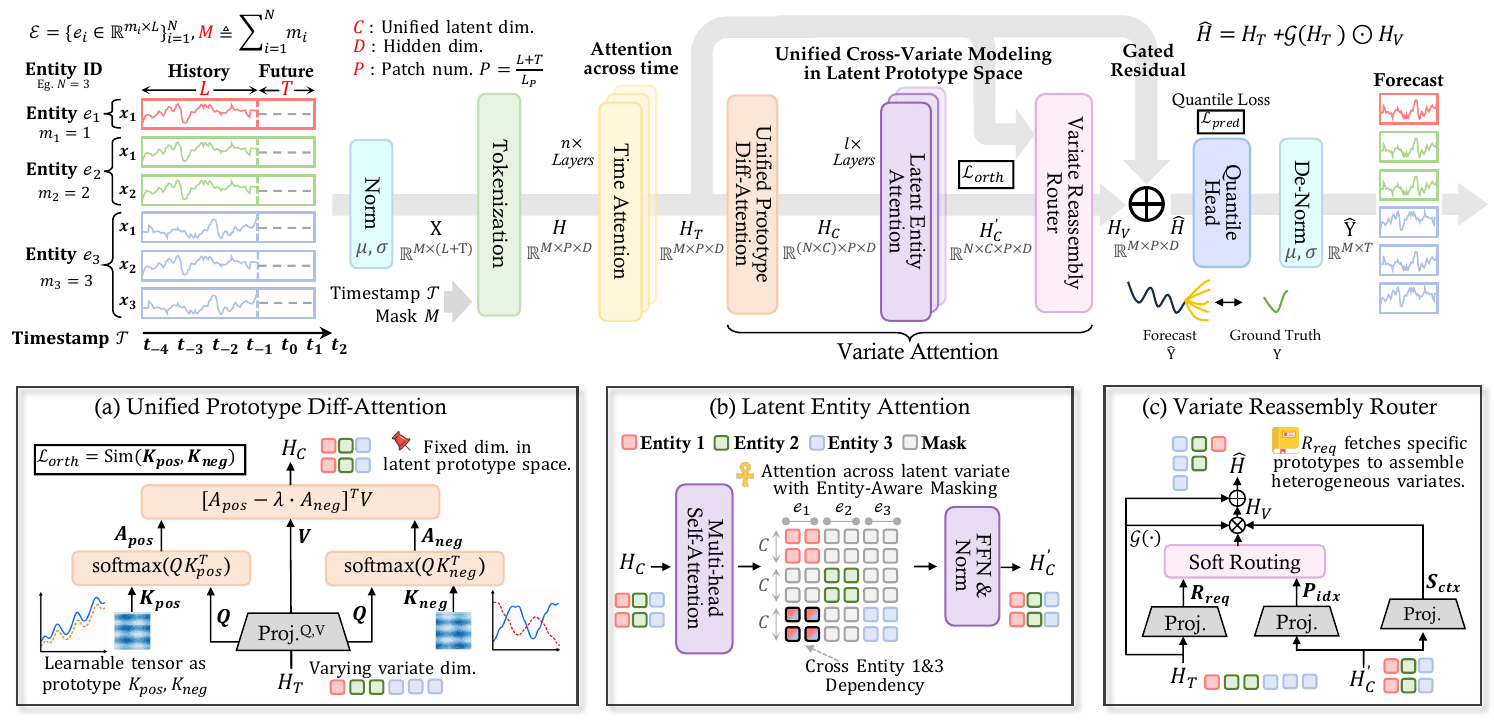}
    \caption{The overall architecture of \NAME. The raw inputs are normalized, tokenized, and processed by Time Attention to extract independent temporal features. The Unified Prototype Diff-Attention (UPDA) then projects these features into a shared prototype space, enabling Latent Entity Attention (LEA) to capture global cross-variate dependencies explicitly. The Variate Reassembly Router (VRR) then dynamically reconstructs variate-specific representations. These are fused with the temporal context and fed into a quantile head to generate the final probabilistic forecasts.}
        \label{fig:pipeline}
\end{figure}

In this section, we present the architecture of \NAME. 
As shown in Figure \ref{fig:pipeline}, \NAME consists of four parts: pre-process of Normalization and Tokenization, Time Attention, Variate Attention and Forecasting Head.
To maintain a concise exposition of the mathematical formulations, we present extended discussions of our \textit{underlying design philosophies} in Appendix \ref{app:method}.

\subsection{Problem Formulation}
\label{sec:problem}

Let $\mathcal{E}=\{\mathbf{e}_i\in\mathbb{R}^{m_i\times L}\}_{i=1}^N$ be a collection of $N$ entities, where each entity $\mathbf{e}_i$ represents a multivariate time series with a specific variate dimensionality $m_i$ and a historical look-back window $L$. A key challenge in TSFM is the heterogeneity of these dimensions, with $m_i$ varying significantly across diverse entities and datasets.
We define the total aggregated dimensionality $M$ across all entities as $\sum_{i=1}^N m_i$. 
The objective of \NAME is to learn a dimension-agnostic mapping $\mathcal{F}_\theta$, parameterized by $\theta$, that transforms the heterogeneous input space into the target predictive space:
\begin{equation}
    \hat{\mathbf{Y}} = \mathcal{F}_\theta(\mathbf{X}), \quad \text{where } \mathbf{X} \in \mathbb{R}^{M \times L}, \hat{\mathbf{Y}} \in \mathbb{R}^{M \times T}.
\end{equation}

\subsection{Normalization and Tokenization}

{\bfseries Normalization.}
We formulate the forecasting task as a unified masked reconstruction paradigm. Given the raw input $\mathbf{X} \in \mathbb{R}^{M \times (L+T)}$ spanning the historical window $L$ and future horizon $T$, we first replace the target future steps with placeholder tokens. To ensure scale-invariance across diverse domains, we apply an arcsine transformation: 
\begin{equation}
\mathbf{\hat{X}} = \arcsin\left(\frac{\mathbf{X} - \mu}{\sigma}\right).
\end{equation}
Crucially, instead of zero-filling or truncating sequences at missing positions \citep{xiaoming2025time}, the instance-wise mean $\mu$ and standard deviation $\sigma$ are computed exclusively from observed values, preserving missing entries for explicit downstream modeling.
% Crucially, instead of zero-filling or truncating sequences at missing positions \citep{xiaoming2025time}, we preserve missing values throughout the pipeline and model them explicitly. Thus, the instance-wise mean $\mu$ and standard deviation $\sigma$ are computed exclusively from the observed values.

{\bfseries Tokenization.}
% To obtain robust and informative representations, we construct an augmented sequence by concatenating the normalized signals $\mathbf{\hat{X}}$ with a relative temporal covariate $\mathcal{T}$ and a binary mask indicator $\mathcal{M}$. Rather than relying on absolute timestamps, which struggle to generalize across heterogeneous sampling frequencies, we employ a normalized relative time encoding to inject a unified sequential ordering. In particular, $\mathcal{T}$ anchors the first forecasting step at $0$, with historical steps linearly decreasing and future steps increasing, all scaled universally by the total context length.
To obtain robust representations, we augment $\mathbf{\hat{X}}$ by concatenating it with a relative timestamps $\mathcal{T}$ and a binary observation mask $\mathcal{M}$. To generalize across heterogeneous sampling frequencies, $\mathcal{T}$ injects a normalized sequential ordering anchored at 0 for the first forecasting step:
\begin{equation}
    \mathcal{T}=\{-\frac{L}{L+T}, \dots, 0, \dots, \frac{T-1}{L+T}\}.
\end{equation}
% Moreover, concatenating the mask $\mathcal{M}$ explicitly enables the model to dynamically distinguish genuine observations from missing entries or masked future steps, significantly improving robustness in real-world deployment scenarios. Subsequently, we partition the augmented sequence into $P=\frac{L+T}{L_p}$ non-overlapping patches of length $L_p$ and project them into the hidden dimension $D$ via a residual patch embedding mechanism:
Meanwhile, the explicit inclusion of $\mathcal{M}$ enables the model to dynamically distinguish genuine observations from missing entries or masked future targets. 
Subsequently, the augmented sequence is partitioned into $P = \frac{L+T}{L_p}$ non-overlapping patches of length $L_p$ and projected into the hidden dimension $D$ via a residual patch embedding mechanism:
\begin{equation}
\mathbf{H} = \text{ResPatchEmbed}(\text{Concat}(\mathbf{\hat{X}}, \mathcal{T}, \mathcal{M}))\in \mathbb{R}^{M \times P \times D}.
\end{equation}
Unlike standard linear projections, the residual patch embedding is designed to seamlessly integrate linear properties with the complex non-linear semantics extracted from the augmented inputs.

\subsection{Time Attention}

To capture the intrinsic evolutionary patterns of each individual variate, \NAME utilizes an encoder-only Transformer architecture (see Appendix \ref{app:TimeAttention}).
Specifically, the Time Attention module consists of $n$ identical encoder layers, each applied independently along the time dimension $D$ for all $M$ variates. Let $\mathbf{H}^{(0)} = \mathbf{H}$ be the input from the Tokenization layer. For each layer $i \in \{1, \dots, n\}$, the hidden state $\mathbf{H}^{(i)}$ is computed as follows:
\begin{equation}
\mathbf{H}^{(i)} = \text{LayerNorm}\left(\mathbf{H}^{(i-1)} + \text{MHA}\left(\mathbf{H}^{(i-1)}\right)\right),
\end{equation}
where MHA denotes the multi-head attention mechanism~\citep{vaswani2017attention}. We apply LayerNorm~\citep{layernorm} to every layer to stabilize the internal activations and facilitate the training of deep foundation architectures. After $n$ successive transformations, the final output is denoted as $\mathbf{H}^{(n)} = \mathbf{H}_T \in \mathbb{R}^{M \times P \times D}$, encapsulating the comprehensive temporal dynamics of each variate.

\subsection{Variate Attention}

A core requirement of time-series foundation models is \textbf{\textit{the ability to transcend rigid, dataset-specific dimensional constraints and learn a unified representation of dependencies across multivariat series}}. 
However, treating different physical variates homogeneously or simply concatenating them leads to severe semantic misalignment. To overcome this, \NAME introduces a unified latent space paradigm. 
As shown in Figure \ref{fig:pipeline}(a–c), these modules progressively aligns the heterogeneous patch embeddings $\mathbf{H}_T$ into a shared prototype space, models both intra- and cross-dataset dependencies, and dynamically reassembles the global context back to the original variate dimensions.

\subsubsection{Unified Prototype Diff-Attention}

To overcome the semantic misalignment inherent in raw-space interactions (detailed discussion in Appendix \ref{app:UPDA}), \NAME projects the heterogeneous temporal embeddings $H_{T}$ into a fixed-$C$-dimensional latent prototype space. We employ a Differential Attention mechanism to explicitly capture both positive and negative semantic affinities.

Concretely, we define two globally shared learnable parameter matrices, $\mathbf{K}_\text{pos}, \mathbf{K}_\text{neg} \in \mathbb{R}^{C \times D}$, representing the synergistic and antagonistic temporal prototypes, respectively. For each entity $\textbf{e}_i$, given its temporal embeddings $\mathbf{h}_T^i \in \mathbb{R}^{m_i \times P \times D}$, we generate the query $\mathbf{Q}^i$ and value $\mathbf{V}^i$ via linear projections. The dual-dependency attention maps, quantifying the affinity between the $m_i$ heterogeneous variates of the $i$-th entity and the $C$ unified prototypes, are computed as:
\begin{equation}
\mathbf{A}_\text{pos}^i = \text{softmax}\left(\frac{\mathbf{Q}^i\mathbf{K}_\text{pos}^\top}{\sqrt{D}}\right), \quad
\mathbf{A}_\text{neg}^i = \text{softmax}\left(\frac{\mathbf{Q}^i\mathbf{K}_\text{neg}^\top}{\sqrt{D}}\right),
\end{equation}
where $\mathbf{A}_\text{pos}^i, \mathbf{A}_\text{neg}^i \in \mathbb{R}^{m_i \times P \times C}$. The unified representation $\mathbf{h}_C^i \in \mathbb{R}^{C \times P \times D}$ for entity $e_i$ is then derived by aggregating the value features based on the differential attention score:
\begin{equation}
\mathbf{h}_C^i = \left[\mathbf{A}_\text{pos}^i - \lambda \cdot \mathbf{A}_\text{neg}^i\right]^\top \mathbf{V}^i,
\end{equation}
where $\lambda$ is a learnable scaling factor and $\mathbf{V}^i=\text{Linear}(\mathbf{h}_T^i)$. 
Notably, while this projection is logically defined at the individual entity level, we implement it concurrently across all $N$ entities to maximize computational efficiency. 
By employing an entity-aware masking strategy, we completely bypass inefficient explicit loops, seamlessly and parallelly transforming the heterogeneous inputs into a perfectly unified latent space $\mathbf{H}_C \in \mathbb{R}^{N \times C \times P \times D}$. To ensure semantic distinctiveness, we apply an orthogonality loss $\mathcal{L}_\text{orth} = \text{Sim}(\mathbf{K}_\text{pos}, \mathbf{K}_\text{neg})$ to constrain the relationship between positive and negative prototypes, where $\text{Sim}(\cdot,\cdot)$ denotes the normalized cosine similarity.

\subsubsection{Latent Entity Attention}

With all representations $\mathbf{H}_C\in \mathbb{R}^{(N \times C) \times P \times D}$ aligned into a shared, dimension-agnostic semantic space, Latent Entity Attention naturally facilitates cross-learning, leveraging shared structural patterns across entirely different domains (see Appendix \ref{app:LEA}).
To be specific, we treat the combined entity and prototype dimensions as the spatial sequence for interaction, and then apply $l$ layers of the standard MHA mechanism to capture the global cross-variate dependencies:
\begin{equation}
\mathbf{H}_C' = \text{LayerNorm}\left(\mathbf{H}_C + \text{MHA}(\mathbf{H}_C)\right),
\end{equation}
where $\mathbf{H}_C' \in \mathbb{R}^{N \times C \times P \times D}$ denotes the refined global context matrix. 
Similar to the Time Attention, this straightforward yet highly effective operation utilizes residual connections and layer normalization to ensure stable representations. By allowing all aligned entities to interact fully within this latent space, $\mathbf{H}_C'$ successfully captures the holistic dynamics necessary for accurate forecasting.

\subsubsection{Variate Reassembly Router}

To accurately reconstruct variate-specific trajectories, \NAME orchestrates a targeted retrieval from the unified prototype space back to individual physical dimensions ($m_i$) via a \textbf{\textit{request-and-dispatch}} mechanism (see Appendix \ref{app:VRR}).
Formally, for each entity $\textbf{e}_i$, we generate the routing components through independent linear projections based on their respective source tensors:
\begin{equation}
\mathbf{R}_\text{req}^i = \text{Linear}(\mathbf{h}_T^i), \quad \mathbf{P}_\text{idx}^i = \text{Linear}(\mathbf{h}_C^{\prime i}), \quad \mathbf{S}_\text{ctx}^i = \text{Linear}(\mathbf{h}_C^{\prime i}).
\end{equation}

Here, the Routing Request ($\textbf{R}_{req}^i$) acts as an \textit{entity identity tag} conveying the unique temporal trajectory of the original variate. It is matched against the Prototype Index ($\textbf{P}_{idx}^i$), an addressable map of the global prototype library, to selectively retrieve refined semantic payloads from the Source Context ($\textbf{S}_{ctx}^i$).
Rather than performing dense token-level interaction, the reconstruction is then executed via a scaled dot-product soft-routing operation, where each variate dynamically allocates its representation across a compact set of latent prototypes:
% The reconstruction is then executed via a scaled dot-product soft-routing operation:
\begin{equation}
\mathbf{h}_V^i =
\text{Route}(\mathbf{R}_{req}^i,\mathbf{P}_{idx}^i)\mathbf{S}_{ctx}^i = \text{softmax}\left(\frac{\mathbf{R}_\text{req}^i (\mathbf{P}_\text{idx}^i)^\top}{\sqrt{D}}\right) \mathbf{S}_\text{ctx}^i.
\end{equation}

This retrieval paradigm successfully reconstructs variate-specific $\mathbf{h}_V^i \in \mathbb{R}^{m_i \times P \times D}$ with high fidelity, smoothly restoring the physical dimensionality to yield $\textbf{H}_V \in \mathbb{R}^{M \times P \times D}$.
Similar to the initial prototype projection, we apply an entity-aware masking strategy during routing, enabling concurrent soft routing across all $N$ entities without explicit loops.

Finally, to maintain robust cross-dataset performance regardless of varying dependency strengths, we introduce an explicit gated residual connection to dynamically fuse the temporal embeddings $\mathbf{H}_T$ with the cross-variate representations $\mathbf{H}_V$. 
The final output $\mathbf{\hat{H}} \in \mathbb{R}^{M \times P \times D}$ is computed as:
\begin{equation}
    \mathbf{\hat{H}} = \mathbf{H}_T + \mathcal{G}(\mathbf{H}_T) \odot \mathbf{H}_V,
\end{equation}
where $\mathcal{G}(\cdot)$ is a gating mechanism with a linear projection followed by a sigmoid activation, and $\odot$ is element-wise multiplication. Thus, \NAME effectively prevents semantic interference in weakly correlated systems while making full use of cross-variate dependencies in strongly correlated ones.

\subsection{Forecasting Head}

Following Chronos-2~\citep{ansari2025chronos2} and Timer-S1~\citep{liu2026timers1}, \NAME adopts a probabilistic forecasting paradigm, predicting future distributions instead of deterministic point estimates. Given the reconstructed representation $\mathbf{\hat{H}}$, we extract the embeddings corresponding to the masked future horizon and apply a linear projection to generate forecasts across a predefined set of quantiles $\mathcal{Q}$. The model is end-to-end optimized using the standard Quantile Loss:
\begin{equation}
\mathcal{L}_\text{pred} = \frac{1}{|\mathcal{Q}| \cdot M \cdot T} \sum_{q \in \mathcal{Q}} \sum_{i=1}^M \sum_{t=1}^T \max\left(q (\mathbf{Y}{i,t} - \mathbf{\hat{Y}}{i,t}^{(q)}), (q - 1) (\mathbf{Y}{i,t} - \mathbf{\hat{Y}}{i,t}^{(q)})\right),
\end{equation}
where $\mathbf{Y}_{i,t}$ represents the ground truth and $\mathbf{\hat{Y}}_{i,t}^{(q)}$ denotes the model's prediction at the $q$-th quantile. The overall training objective combines this with the prototype orthogonality loss: $\mathcal{L} = \mathcal{L}_\text{pred} + \alpha \mathcal{L}_\text{orth}$, where $\alpha$ is a hyper-parameter balancing the forecasting and orthogonality objectives.

During inference, the predictions are mapped back to their original physical scale $\mathbf{\tilde{Y}}^{(q)}$ via a straightforward de-normalization process, applying the sine transformation followed by De-Norm using the preserved instance-wise statistics.
\begin{equation}
\mathbf{\tilde{Y}}^{(q)} = \sigma \cdot \sin(\mathbf{\hat{Y}}^{(q)}) + \mu.
\end{equation}

%% file: sections/4-training_details.tex
\section{Training Details}
\subsection{Pre-Training Corpus}
Our pre-training corpus combines large-scale real-world and synthetic time series data spanning several domains. In addition to public datasets from \textsc{GIFT-Eval}~\citep{aksu2024gift} , \textsc{Chronos}~\citep{ansari2024chronos} and \textsc{QuitoBench}~\citep{xue2026quitobench}, it includes synthetic univariate and multivariate series designed to increase diversity in temporal patterns and dependency structures.
Specifically, synthetic univariate data are generated through data mixing and stochastic process sampling, while multivariate series are constructed by grouping related univariate signals and injecting explicit cross-variate dependencies, including both instantaneous and temporal interactions. The details can be found in Appendix \ref{app:pre-training corpus}.

\subsection{Training Infrastructure and Config}

To enable scalable pretraining on massive, heterogeneous time-series corpora, we build \NAME upon Megatron-LM~\citep{megatron-lm} and design a custom sampling pipeline to balance data distribution across diverse domains. Furthermore, to address the heterogeneity in variate dimensionality, we implement a runtime multivariate sampling strategy that dynamically balances the number of consuming variates per batch, thereby improving GPU utilization and training stability. Further implementation details are provided in Appendix \ref{app:sampling details}.

\NAME features a hidden dimension of $D=1024$, a patch length of $L_p=16$, and utilizes $n=16$ Time Attention layers alongside $l=16$ Entity Attention layers ($16$ heads per layer). By accommodating up to $512$ input tokens and $30$ output tokens, it achieves a maximum context length of $L=8192$ and a prediction length of $T=480$ in a single inference pass. The model is pre-trained on a cluster of NVIDIA B200-180GB GPUs for one million iterations with a global batch size of $384$ using \texttt{bf16} precision, optimized by a joint quantile and orthogonality loss. 
We adopt the AdamW optimizer~\citep{loshchilov2018adamw} with $\beta_1=0.9$, $\beta_2=0.95$, and a weight decay of $0.1$. 
The learning rate warms up linearly to $6 \times 10^{-5}$ over the first $0.1\%$ of steps, followed by a cosine decay to $6 \times 10^{-6}$.
As shown in Figure \ref{fig:train_iter}, the training process is highly stable, with the loss curve converging smoothly and robustly.

%% file: sections/5-experiment.tex
\section{Experiments}

\begin{figure}[!tb]
    \centering
    \includegraphics[width=\textwidth]
    {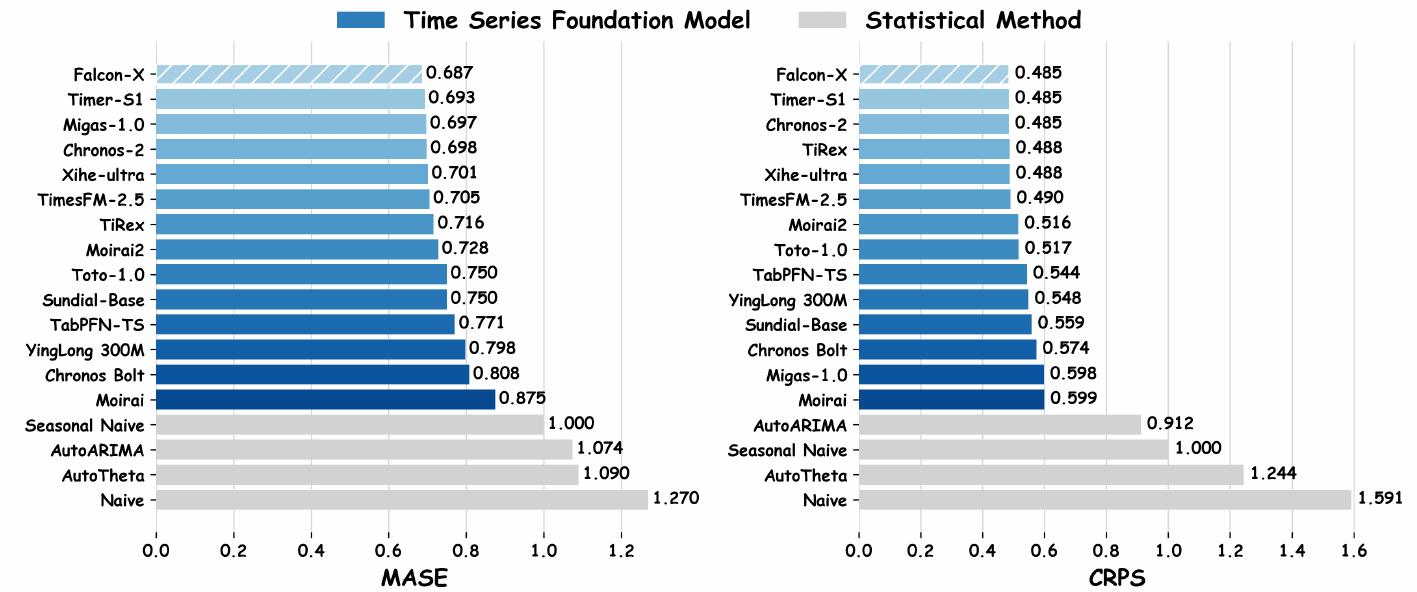}
    \caption{Performance of \NAME on the GIFT-Eval leaderboard.}
    \label{fig:mainresults_gift}
\end{figure}

\begin{figure}[!t]
    \centering
    \includegraphics[width=1\textwidth]{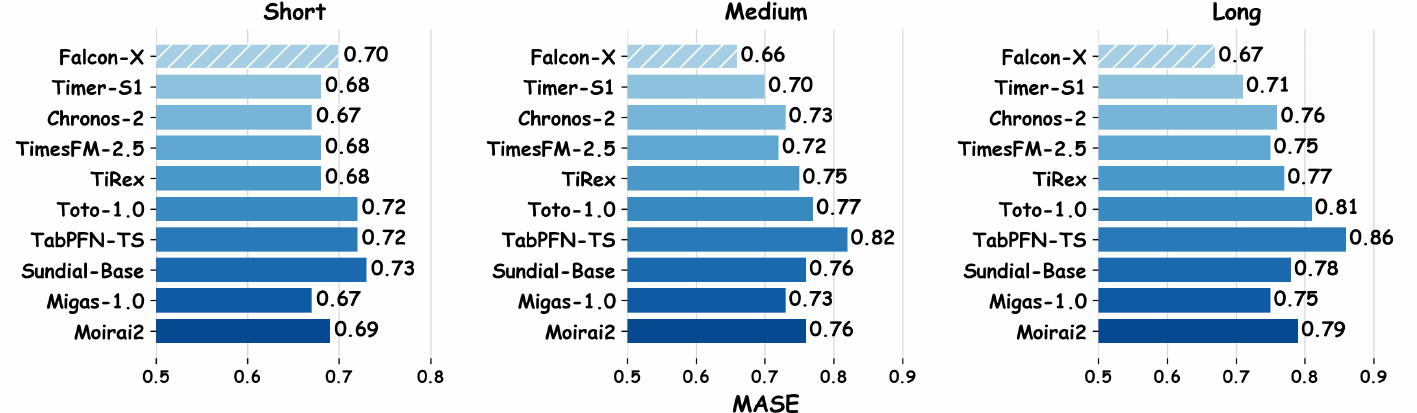}
    \caption{Performance (MASE) of \NAME on the GIFT-Eval leaderboard, grouped by the term length. \NAME exhibits remarkable stability across all horizons.}
    \label{fig:term_gift}
\end{figure}

\subsection{Main Results}
\label{sec:mainresults}

We evaluate Falcon-X on two comprehensive benchmarks, GIFT-Eval~\citep{aksu2024gift} and fev-bench~\citep{shchur2025fev}, using MASE and CRPS to measure point forecasting accuracy and probabilistic calibration, respectively. As shown in Figure \ref{fig:mainresults_gift}, Falcon-X achieves the best MASE and tied-best CRPS on GIFT-Eval, reaching $0.687$ MASE and $0.485$ CRPS. Compared with strong competing time-series foundation models, Falcon-X delivers lower or comparable errors: it improves over Timer-S1~\citep{liu2026timers1} by $0.9\%$ in MASE, over Chronos-2~\citep{ansari2025chronos2} by $1.6\%$ in MASE, over TiRex~\citep{auertirex} by $4.1\%$ in MASE, over Moirai 2.0~\citep{liu2025moirai} by $5.6\%$ in MASE, and over Toto-1.0~\citep{cohen2025toto} by $8.4\%$ in MASE. For probabilistic forecasting, Falcon-X matches Timer-S1~\citep{liu2026timers1} and Chronos-2~\citep{ansari2025chronos2} at $0.485$ CRPS under the reported precision, while improving over TiRex~\citep{auertirex} by $0.6\%$, over Moirai 2.0~\citep{liu2025moirai} by $6.0\%$, and over Toto-1.0~\citep{cohen2025toto} by $6.2\%$ in CRPS. These results demonstrate that explicit latent prototype alignment provides an effective and competitive mechanism for heterogeneous multivariate forecasting, especially when compared with raw-space variate mixing used by representative multivariate TSFMs.

We further analyze the robustness of Falcon-X across different prediction horizons on GIFT-Eval. Figure \ref{fig:term_gift} reports the MASE results grouped into short-, medium-, and long-term forecasting settings. Falcon-X obtains $0.70$ MASE in the short-term setting, remaining competitive although several baselines achieve lower short-horizon errors, such as Chronos-2~\citep{ansari2025chronos2} with $0.67$ MASE and Timer-S1~\citep{liu2026timers1} and TiRex~\citep{auertirex} with 0.68 MASE. More importantly, its advantage becomes clear as the forecasting horizon increases: Falcon-X achieves the best medium-term and long-term results, with $0.66$ and $0.67$ MASE, respectively. In contrast, Chronos-2~\citep{ansari2025chronos2} increases from $0.67$ in the short-term setting to $0.73$ and $0.76$ in the medium- and long-term settings, while TabPFN-TS~\citep{hoo2025tables} degrades from $0.72$ to $0.82$ and $0.86$. These results indicate that Falcon-X is particularly effective under extended horizons, suggesting that the latent prototype routing mechanism can capture durable cross-variate dynamics and mitigate horizon-wise error accumulation.

\begin{figure}[!tb]
    \centering
    \includegraphics[width=1\textwidth]{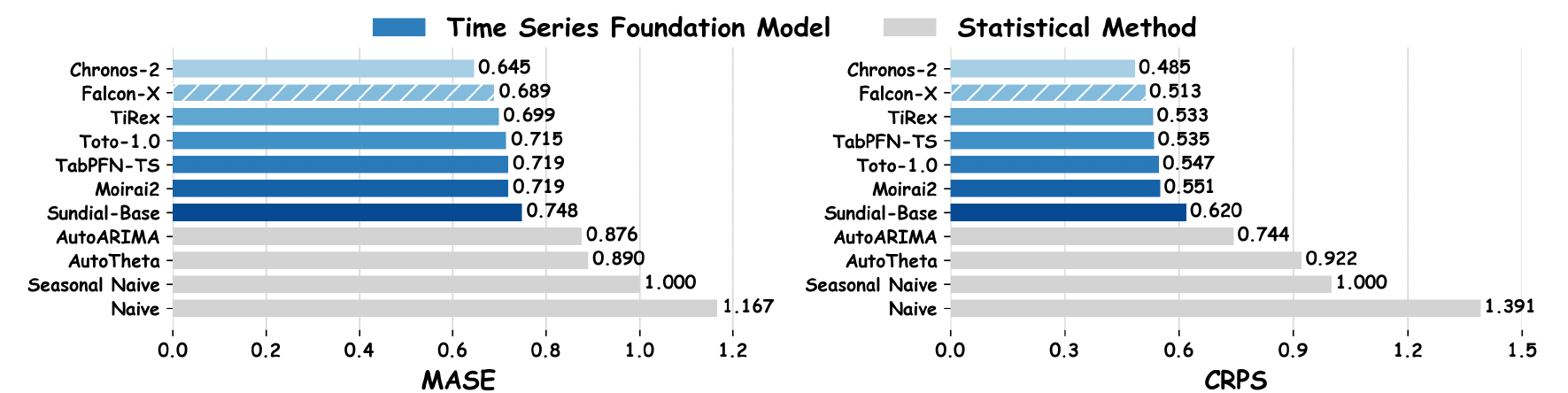}
    \caption{Performance of \NAME on the fev-bench leaderboard.}
        \label{fig:mainresults_fev}
\end{figure}

On fev-bench, Falcon-X also exhibits strong generalization performance, as shown in Figure \ref{fig:mainresults_fev}. Falcon-X ranks second behind Chronos-2~\citep{ansari2025chronos2}, achieving $0.689$ MASE and $0.513$ CRPS, compared with Chronos-2's $0.645$ MASE and $0.485$ CRPS. The relative gap to Chronos-2 is $6.8\%$ on MASE and $5.8\%$ on CRPS, while Falcon-X relies strictly on endogenous target series rather than additional past-only or future-known covariates. Beyond Chronos-2~\citep{ansari2025chronos2}, Falcon-X consistently outperforms other recent foundation models, including TiRex~\citep{auertirex}, Toto-1.0~\citep{cohen2025toto}, TabPFN-TS~\citep{hoo2025tables}, Moirai 2.0~\citep{liu2025moirai}, and so on. For example, Falcon-X improves over TiRex~\citep{auertirex} by $1.4\%$ in MASE and $3.8\%$ in CRPS, and over Moirai 2.0~\citep{liu2025moirai} by $4.2\%$ in MASE and $6.9\%$ in CRPS. These results confirm that the proposed dual-dependency architecture provides strong relational expressivity and structural adaptability across diverse real-world forecasting tasks.

\subsection{Ablation Studies}
\label{sec:ablation}
\begin{figure}[!t]
    \centering
    \includegraphics[width=\textwidth]{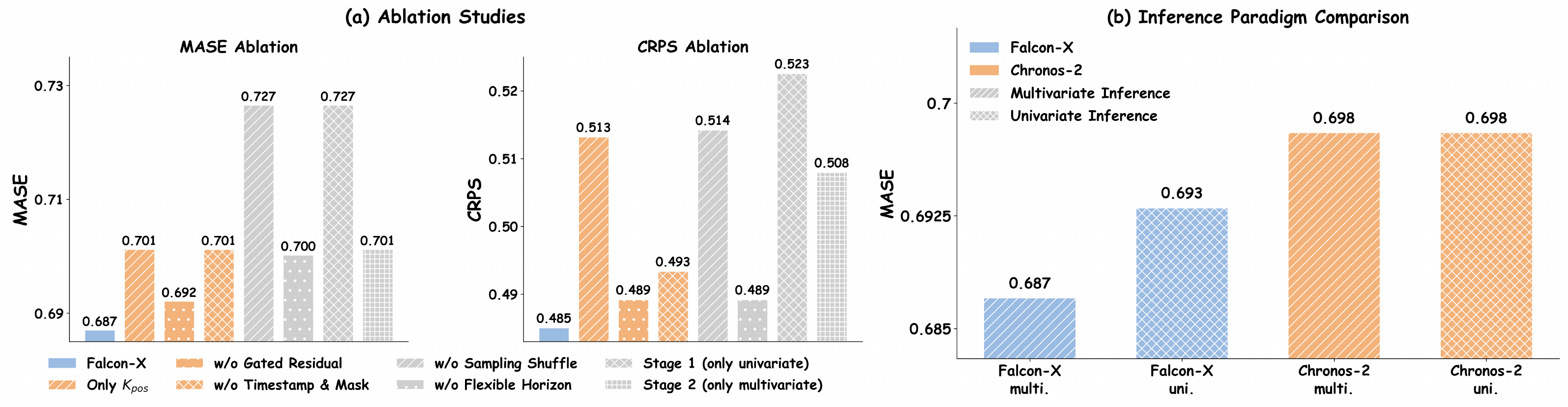}
    \caption{%  Comprehensive analysis of \NAME. 
    (a) Ablation studies validating the necessity of key architectural components and training strategies. (b) Inference paradigm comparison, highlighting our robust multivariate modeling against Chronos-2.}
    \label{fig:analysis}
\end{figure}

As shown in Figure \ref{fig:analysis}(a), to isolate the contributions of the architecture and optimization pipeline, we conduct comprehensive ablation studies on both model components and training strategies.

{\bfseries Module Ablation.}
We first evaluate the structural designs of \NAME. 
(i) \textbf{Only $\textbf{K}_\text{pos}$:} Removing the negative prototype key ($\textbf{K}_\text{neg}$) causes the most severe performance drop, verifying that modeling negative affinities is essential for heterogeneous series interactions. 
(ii) \textbf{w/o gated residual:} Removing the gated residual connection degrades performance, showing its importance in context-aware filtering cross-variate information in various datasets. 
(iii) \textbf{w/o timestamp \& mask:} Excluding the relative timestamp $\mathcal{T}$ and mask $\mathcal{M}$ reduces robustness to irregular sampling and missing values.

{\bfseries Strategy Ablation.}
We further analyze the impact of our data processing and training pipeline. 
(i) \textbf{w/o sampling shuffle:} Disabling variate shuffling significantly hurts performance, indicating that random permutation is crucial for learning content-driven rather than index-dependent relationships. 
(ii) \textbf{w/o flexible horizon:} Replacing flexible horizon sampling with fixed-length prediction weakens generalization across unseen forecasting horizons. 
(iii) \textbf{Two-stage vs. Joint training:} We compare direct joint training with a two-stage curriculum consisting of: (Stage 1) univariate pre-training for temporal modeling initialization, and (Stage 2) multivariate fine-tuning with univariate replay. Joint training consistently performs better, indicating that \NAME can naturally unify temporal dynamics and cross-variate interactions within a single optimization process, without relying on carefully staged training curricula.

\begin{figure}[!t]
    \centering
    \includegraphics[width=\textwidth]{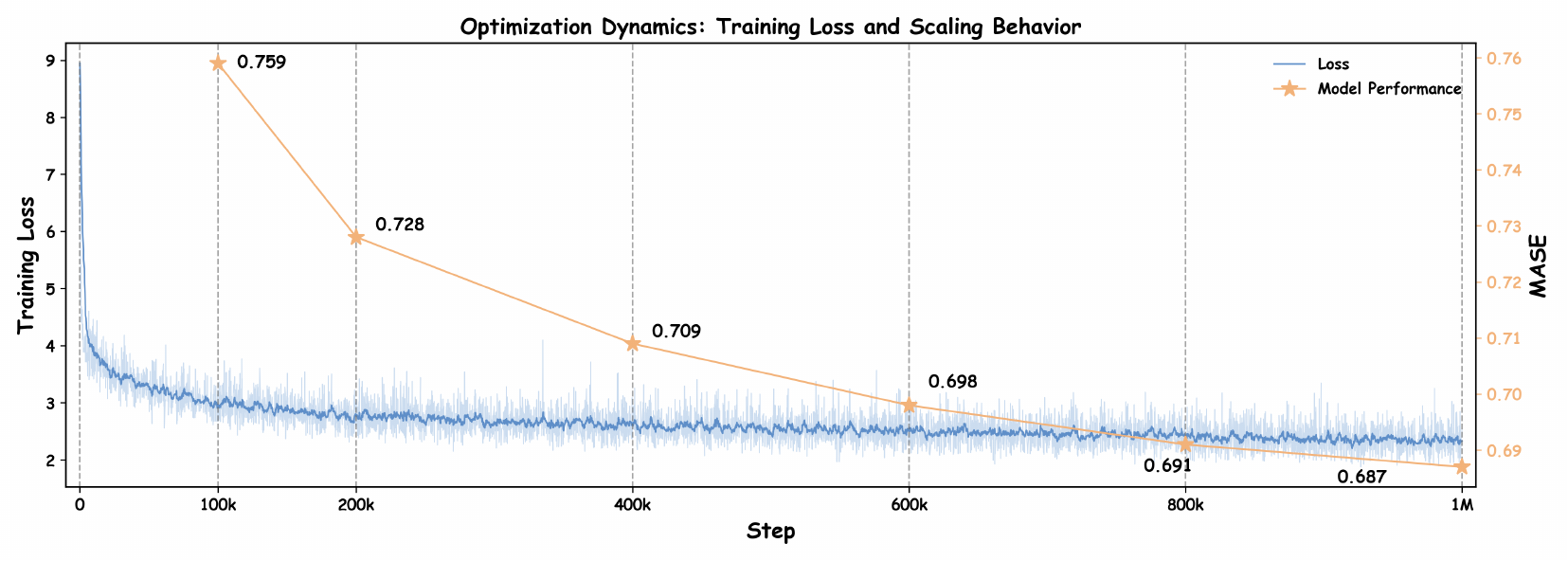}
    \caption{Stable training dynamics and performance scaling over one million iterations.}
    \label{fig:train_iter}
\end{figure}

\subsection{Inference Setting Analysis}
\label{sec:inference}
We compare inference paradigms against Chronos-2 in Figure \ref{fig:analysis}(b). 
On GIFT-Eval, Chronos-2~\citep{ansari2025chronos2} exhibits nearly identical performance regardless of whether its group attention is enabled, indicating that raw-space variate mixing contributes little effective relational information. This reveals a severe semantic collapse, where cross-variate interaction degenerates into univariate-like behavior.
% On GIFT-Eval, Chronos-2’s performance remains identical whether its group attention is enabled or not. This exposes a severe semantic collapse, where raw-space mixing fails to capture dependencies and degenerates into univariate-like behavior. 
% In contrast, enabling multivariate inference in \NAME consistently improves accuracy on multivariate tasks over its univariate inference mode. 
In contrast, enabling multivariate inference in \NAME consistently improves accuracy on multivariate tasks over its univariate inference mode, demonstrating that \NAME can effectively capture transferable cross-variate dependencies.
Crucially, this cross-variate enhancement fully preserves the accuracy on univariate forecasting, proving that our latent routing successfully extracts global synergistic context without corrupting individual temporal signals.

\subsection{Influence of Key Parameter}
\label{sec:sentivity}
We analyze the sensitivity of \NAME to two key architectural parameters.

\begin{figure}[!t]
    \centering
    \includegraphics[width=\textwidth]{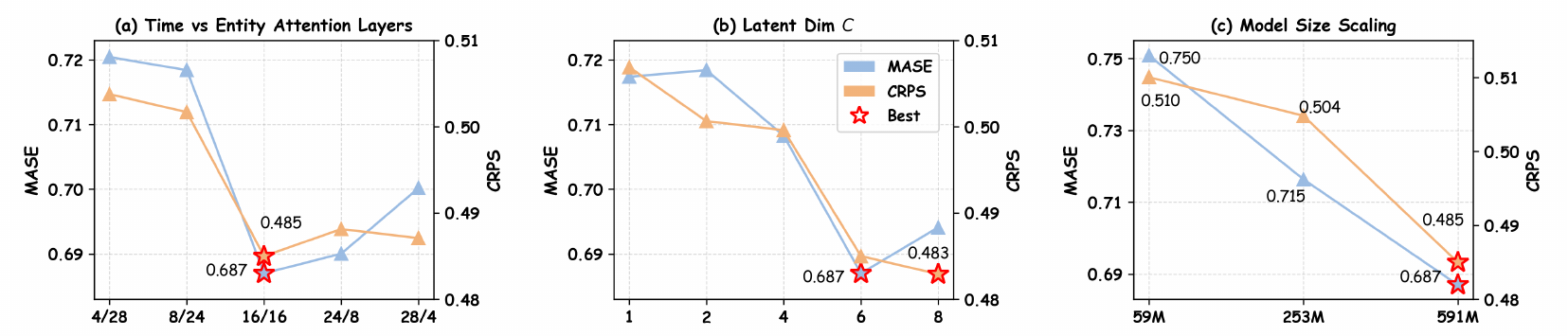}
    \caption{Sensitivity and scaling analysis. (a) Performance impact of layer allocation between Time Attention $n$ and Entity Attention $l$. (b) Sensitivity to the latent prototype dimension $C$. (c) Consistent performance scaling across increasing model parameter sizes (from 59M to 591M).}
        \label{fig:sentivity}
\end{figure}

{\bfseries Depth Distribution ($n$ vs. $l$).}
We investigate the layer allocation between Time Attention ($n$) and Latent Entity Attention ($l$) under a fixed depth budget. As shown in Figure \ref{fig:sentivity}(a), allocating sufficient capacity to temporal modeling is essential, while excessive cross-variate routing significantly degrades performance. 
This indicates that robust temporal modeling is the foundation of forecasting, while cross-variate interaction provides complementary gains.
\NAME achieves the best trade-off with a balanced $16/16$ configuration.

{\bfseries Latent Prototype Dimension ($C$).} 
We evaluate the representational capacity of the unified semantic space by varying $C$. Figure \ref{fig:sentivity}(b) demonstrates that a severely restricted dimension ($C \le 2$) induces an information bottleneck, leading to semantic over-compression. Conversely, expanding the dimension enhances relational expressivity. The model achieves peak accuracy at $C=6$ and $C=8$, striking a perfect balance between capturing diverse dynamics and avoiding redundant noise.

\subsection{Scaling Analysis}
\label{sec:scaling}
We evaluate the scaling behaviors of \NAME across training iterations and parameter sizes. As shown in Figure \ref{fig:train_iter}, the training dynamics exhibit a smooth, stable loss descent alongside continuous forecasting performance gains throughout the entire $1 \times 10^6$ steps. 
Furthermore, scaling the model capacity from 59M ($l=n=8,D=512$) to 253M ($l=n=12,D=768$) and up to 591M ($l=n=16,D=1024$) yields strictly predictable improvements in both MASE and CRPS (Figure \ref{fig:sentivity}(c)). These consistent trajectories demonstrate that our decoupled architecture strictly adheres to neural scaling laws, confirming its robust scalability and vast capacity to absorb massive heterogeneous time series without saturation.

\subsection{Case Study}
\label{sec:casestudy}
To qualitatively analyze \NAME, we present representative forecasting cases from GIFT-Eval.

\begin{figure}[!t]
    \centering
    \includegraphics[width=\textwidth]{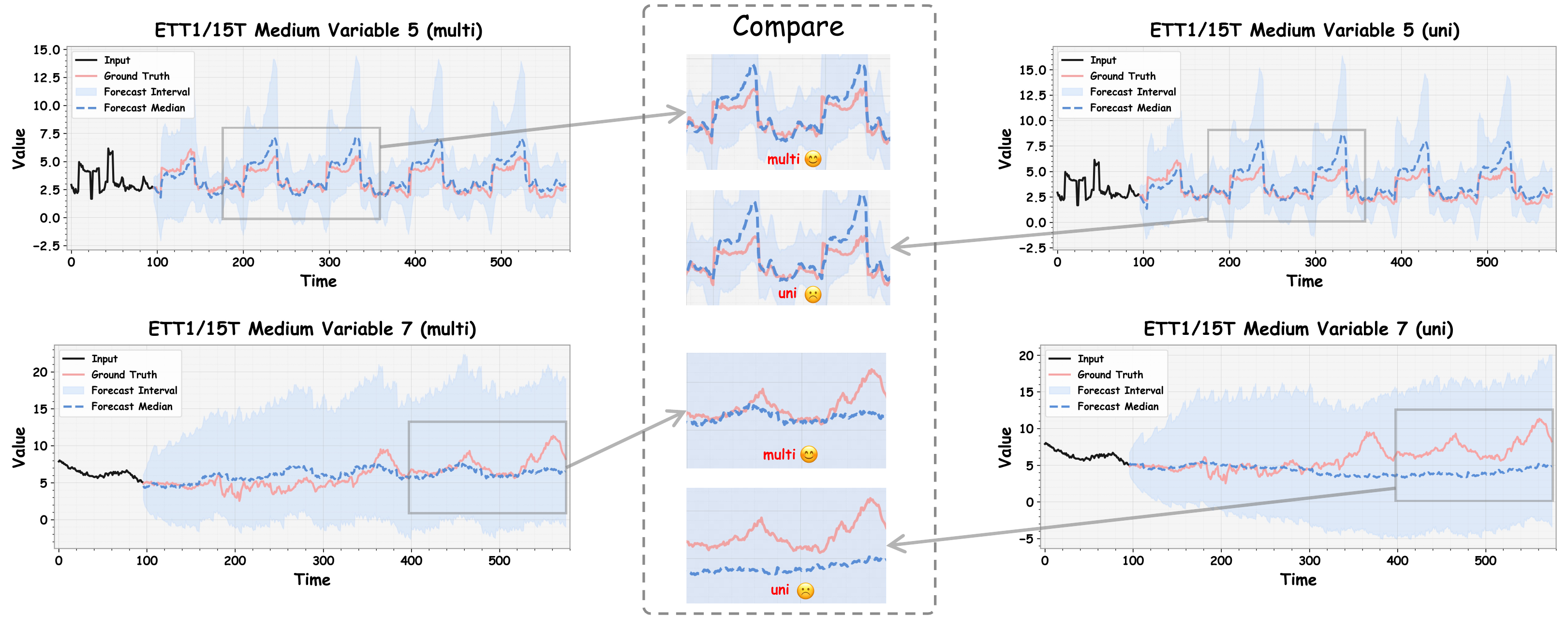}
    \caption{Case study on ETT1/15T dataset. In univariate inference mode, forecasts gradually deviate from the ground truth due to the absence of global context. In contrast, \NAME's multivariate inference effectively leverages cross-variate signals to calibrate trajectories.}
    \label{fig:casestudy}
\end{figure}

{\bfseries Multivariate vs. Univariate Inference.} 
We compare multivariate inference with channel-independent inference on highly correlated sequences from ETT1/15T (see Figure \ref{fig:casestudy}). Without cross-variate interactions, the channel-independent setting gradually drifts from the ground truth under complex temporal shifts. In contrast, \NAME leverages its unified latent prototype space to aggregate complementary signals across variates, producing substantially more accurate trajectories.

{\bfseries Positive and Negative Dependency Modeling.} 
We further examine the ability of \NAME to capture dual dependencies. As shown in Figure \ref{fig:bit_fast_5t}, \NAME accurately models positive correlations with synchronized trends. More importantly, in Figure \ref{fig:biz_application_10s}, it successfully captures negative correlations with opposing dynamics. These results validate the effectiveness of our Unified Prototype Diff-Attention for modeling both synergistic and antagonistic relationships.

%% file: sections/6-conclusion.tex
\section{Conclusion}

In this paper, we identify two fundamental challenges of existing multivariate time series foundation models: semantic alignment and relational expressivity. To address these issues, we propose \NAME, a novel modeling paradigm that decouples physical variables into a unified latent prototype space. 
By introducing the Unified Prototype Diff-Attention, our architecture effectively captures both synergistic and antagonistic correlations. Additionally, a Variate Reassembly Router ensures robust global context fusion across diverse domains. 
Extensive evaluations on GIFT-Eval and fev-bench demonstrate that \NAME achieves superior performance, showcasing exceptional scalability and zero-shot transferability.
We hope that this work will contribute to the development of more unified and expressive foundation models for time series.

%% file: sections/appendix.tex
\section{Methodology Design Philosophies}
\label{app:method}

\subsection{Overall Architecture}
\label{app:OverallArchitecture}

As shown in Figure \ref{fig:pipeline}, the architecture of \NAME follows a hierarchical transformation pipeline that progressively aligns heterogeneous variates into a unified latent space, and ultimately reconstructs them for accurate forecasting.
We formulate the forecasting task as a unified masked reconstruction paradigm. Formally, the input heterogeneous series $\mathbf{X} \in \mathbb{R}^{M \times (L+T)}$ is first subjected to instance normalization and tokenization to generate the time tokens $\mathbf{H} \in \mathbb{R}^{M \times P \times D}$, where $P=\frac{L+T}{L_p}$ is the number of patches and $L_p$ is the length of patches. 
These tokens are then processed through time attention layers, yielding temporal representations $\mathbf{H}_T \in \mathbb{R}^{M \times P \times D}$. 

To bridge the dimensionality gap, \NAME employs the Unified Prototype Diff-Attention (UPDA) to project the disparate $N$ entities into a compact, unified prototype space $\mathbf{H}_C \in \mathbb{R}^{(N \times P) \times C \times D}$. 
Following cross-variate interactions via Latent Entity Attention (LEA), which yields the refined context $\mathbf{H}_C'$, the Variate Reassembly Router (VRR) performs a soft-routing operation to retrieve and reassemble the latent representations back into the entity-specific space $\mathbf{H}_V \in \mathbb{R}^{M \times P \times D}$ by matching the routing request in $\mathbf{H}_T$ with the prototype index in $\mathbf{H}_C'$. 

Finally, the reassembled entities $\mathbf{H}_V$ are fused with the temporal representations $\mathbf{H}_T$ and mapped through a quantile forecasting head to produce the final predictive output $\hat{\mathbf{Y}} \in \mathbb{R}^{M \times T}$, completing the end-to-end flow from raw heterogeneous inputs to unified latent representations and back to structured forecasts.

\subsection{Time Attention}
\label{app:TimeAttention}

To capture the intrinsic evolutionary patterns of each individual variate, \NAME utilizes an encoder-only Transformer architecture.
A critical design choice in \NAME is the deliberate decoupling of temporal and cross-variate modeling. Unlike many existing multivariate models, which interleave temporal and spatial mixing, leading to semantic entanglement whereby a variate's subtle temporal signal is prematurely confounded by the noisy dependencies of its heterogeneous neighbors, \NAME prioritizes establishing a robust temporal module.
Stacking $n$ layers of Time Attention before any cross-variate interaction ensures that the temporal evolutive state of each variate is fully distilled and stabilised. This provides a clean, time-aware foundation for the subsequent Variate Attention process.

\subsection{Variate Attention}
\label{app:VariateAttention}

A core requirement of time-series foundation models is \textbf{\textit{the ability to transcend rigid, dataset-specific dimensional constraints and learn a unified representation of dependencies across multivariat series}}. 
However, treating different physical variates homogeneously or simply concatenating them leads to severe semantic misalignment. To overcome this, \NAME introduces a unified latent space paradigm. 
As shown in Figure \ref{fig:pipeline}(a–c), these modules progressively aligns the heterogeneous patch embeddings $\mathbf{H}_T$ into a shared prototype space, models both intra- and cross-dataset dependencies, and dynamically reassembles the global context back to the original variate dimensions.

\subsubsection{Unified Prototype Diff-Attention}
\label{app:UPDA}

The primary challenge in modeling the foundations of time series lies in reconciling diverse physical variates within a unified semantic space. 
Dataset-level group mixing~\citep{ansari2025chronos2} relies on dense intra-dataset attention, lacking structural abstraction and incurring quadratic complexity $\mathcal{O}(M^2)$.
To address this, \NAME introduces prototype alignment, projecting heterogeneous variates into a fixed set of learnable latent temporal prototypes of dimension $C$.

Instead of relying on arbitrary physical indexing, this paradigm dynamically allocates the full representation of each independent variate across the $C$ universal prototypes based on their intrinsic semantic affinity, thereby achieving explicit semantic unification. 
This explicit mapping resolves the issue of semantic misalignment by aligning variates with similar temporal dynamics to the same semantic anchors, regardless of their original dataset or spatial proximity. Furthermore, by forcing the representations through this fixed-dimensional prototype space, the model inherently performs structural denoising, filtering out localized noise and isolating the most salient temporal patterns.
Importantly, this projection replaces dense intra-dataset attention with cross-attention, decoupling the computational bottleneck from the physical dimensionality. The complexity is thus reduced to a strictly linear $\mathcal{O}(M \cdot C)$, effortlessly accommodating extreme-dimensional modeling since $C \ll M$.

Furthermore, we observe that negative correlations among heterogeneous variates, which are critical for capturing counteracting interactions across diverse time series, are difficult to exploit in standard Transformer architectures. This limitation stems from the non-negative nature of the softmax attention function, which restricts attention scores to the range $[0, 1]$ and thus prevents the explicit modeling of opposing trends. 
To address this, we draw inspiration from differential attention~\citep{diffattn}. While originally proposed for attention noise suppression, \NAME repurposes this mechanism to capture dual-dependency dynamics.
By introducing positive and negative learnable keys, the model explicitly represents both synergistic and antagonistic relationships, yielding improved expressiveness of the cross-variate latent space.

\subsubsection{Latent Entity Attention}
\label{app:LEA}

Following the alignment of heterogeneous variates into the unified prototype space, this module models the comprehensive interactions among different variates. As all representations now reside in a shared, dimension-agnostic semantic space rather than their original disparate physical dimensions, Latent Entity Attention naturally facilitates cross-learning. This enables \NAME to leverage and transfer shared structural patterns across entirely different domains, thereby significantly enhancing zero-shot cross-dataset generalization.

\subsubsection{Variate Reassembly Router}
\label{app:VRR}

After capturing comprehensive dependencies in the unified prototype space, the model must reassemble this global context back into the original heterogeneous dimensions ($m_i$).
\NAME formulates this reassembly as a targeted retrieval from the abstract prototype space to individual variate trajectories. 
The aim is to reconstruct the heterogeneous variates, each with distinct temporal patterns, by retrieving relevant information from the unified latent prototypes. 

This is orchestrated via a \textbf{\textit{request-and-dispatch}} mechanism:
the Routing Request ($\mathbf{R}_\text{req}$), derived from $\mathbf{h}_T^i$, acts as a structural query conveying the specific physical dimensionality and unique temporal trajectory of the original variate, effectively serving as \textit{entity identity tag}.
The request is then matched against the Prototype Index ($\mathbf{P}_\text{idx}$), which is an addressable map of the global prototype library. 
Meanwhile, the Source Context ($\mathbf{S}_\text{ctx}$) delivers the refined semantic payloads. 
Using local, specific trajectories to selectively retrieve unified global prototypes enables this routing paradigm to reconstruct variate-specific patterns with high fidelity.

Finally, considering the significant variance in cross-variate dependencies across diverse datasets, forcing a uniform integration of the global context could introduce detrimental noise to datasets with inherently weak variate correlations. To maintain strict cross-dataset robustness, we introduce an explicit gated residual connection to dynamically fuse the temporal embeddings $\mathbf{H}_T$ with the cross-variate representations $\mathbf{H}_V$. 
Consequently, it effectively prevents semantic interference in weakly correlated systems while making full use of cross-variate dependencies in strongly correlated ones.

\renewcommand{\arraystretch}{0.7}
\begin{table}[htbp]
\centering
\captionsetup{skip=4pt}
\caption{Summary statistics of univariate pre-training datasets.}
\label{tab:univariate_pretrain_dataset}
\begin{threeparttable}
\resizebox{0.88\textwidth}{!}{%
\begin{tabular}{
@{} l l
S[table-format=7.0]
S[table-format=2.0]
S[table-format=11.0]
l l @{}
}
\toprule
\textbf{Dataset Name} &
\textbf{Frequency} &
\multicolumn{1}{c}{\textbf{Time Series}} &
\multicolumn{1}{c}{\textbf{Variates}} &
\multicolumn{1}{c}{\textbf{Time Points}} &
\textbf{Domain} &
\textbf{Source} \\
\midrule
\texttt{BDG-2} & H & 611 & 1 & 9454968 & Energy & GIFT-Eval \\
\texttt{BEIJING\_SUBWAY\_30MIN} & 30T & 276 & 2 & 433872 & Transport & GIFT-Eval \\
\texttt{CIF 2016} & M & 72 & 1 & 6334 & Finance & GIFT-Eval \\
\texttt{CMIP6} & 6H & 270336 & 53 & 1973452800 & Nature & GIFT-Eval \\
\texttt{ERA5} & H & 245760 & 45 & 2146959360 & Nature & GIFT-Eval \\
% \texttt{Electricity} & H, W & 642 & 1 & 8493660 & Energy & Chronos \\
\texttt{HZMETRO} & 15T & 80 & 2 & 190160 & Transport & GIFT-Eval \\
\texttt{LOS\_LOOP} & 5T & 207 & 1 & 7094304 & Transport & GIFT-Eval \\
\texttt{LargeST} & 5T & 42333 & 1 & 4452510528 & Transport & GIFT-Eval \\
\texttt{M1} & A, M, Q & 921 & 1 & 57882 & Finance & GIFT-Eval \\
\texttt{M3} & A, M, Q & 3003 & 1 & 209114 & Finance & GIFT-Eval \\
\texttt{NN5} & D, W & 222 & 1 & 93240 & Finance & GIFT-Eval \\
\texttt{PEMS03} & 5T & 358 & 1 & 9382464 & Transport & GIFT-Eval \\
\texttt{PEMS04} & 5T & 307 & 3 & 5216544 & Transport & GIFT-Eval \\
\texttt{PEMS07} & 5T & 883 & 1 & 24921792 & Transport & GIFT-Eval \\
\texttt{PEMS08} & 5T & 170 & 3 & 3035520 & Transport & GIFT-Eval \\
\texttt{PEMS\_BAY} & 5T & 325 & 1 & 16941600 & Transport & GIFT-Eval \\
\texttt{Q-TRAFFIC} & 15T & 45148 & 1 & 264386688 & Transport & GIFT-Eval \\
\texttt{Quito} & 10T, H & 33806 & 5 & 313269828 & Various & QuitoBench \\
\texttt{Residential Power} & T & 504 & 3 & 271333509 & Energy & GIFT-Eval \\
\texttt{SHMETRO} & 15T & 288 & 2 & 2536992 & Transport & GIFT-Eval \\
% \texttt{Solar} & 5T, H & 10332 & 1 & 588304080 & Energy & Chronos \\
\texttt{Solar} & 5T & 10332 & 1 & 588304080 & Energy & Chronos \\
\texttt{Taxi} & 30T, H & 70412 & 1 & 56793348 & Transport & Chronos \\
\texttt{Tourism} & A, M, Q & 1212 & 1 & 150822 & Finance & GIFT-Eval \\
\texttt{Traffic} & H, W & 1724 & 1 & 15060864 & Transport & GIFT-Eval \\
\texttt{Uber TLC} & D, H & 524 & 1 & 1176531 & Transport & GIFT-Eval \\
\texttt{Weatherbench} & D, H, W & 675840 & 1 & 82753646592 & Nature & Chronos \\
\texttt{Wind Farms} & D, H, T & 1011 & 1 & 175154333 & Energy & Chronos \\
\texttt{alibaba\_cluster\_trace\_2018} & 5T & 58409 & 2 & 95192530 & Web & GIFT-Eval \\
\texttt{australian\_electricity\_demand} & 30T & 5 & 1 & 1153584 & Energy & GIFT-Eval \\
\texttt{azure\_vm\_traces\_2017} & 5T & 159472 & 1 & 885522908 & Web & GIFT-Eval \\
\texttt{beijing\_air\_quality} & H & 12 & 11 & 420768 & Nature & GIFT-Eval \\
\texttt{bitcoin\_with\_missing} & D & 18 & 1 & 81918 & Finance & GIFT-Eval \\
\texttt{borealis} & H & 15 & 1 & 83269 & Energy & GIFT-Eval \\
\texttt{borg\_cluster\_data\_2011} & 5T & 143386 & 2 & 537552854 & Web & GIFT-Eval \\
\texttt{buildings\_900k} & H & 1792328 & 1 & 15702585608 & Energy & GIFT-Eval \\
\texttt{bull} & H & 41 & 1 & 719304 & Energy & GIFT-Eval \\
\texttt{cdc\_fluview\_ilinet} & W & 75 & 5 & 63903 & Healthcare & GIFT-Eval \\
\texttt{cdc\_fluview\_who\_nrevss} & W & 74 & 4 & 41760 & Healthcare & GIFT-Eval \\
\texttt{china\_air\_quality} & H & 437 & 6 & 5739234 & Nature & GIFT-Eval \\
\texttt{cockatoo} & H & 1 & 1 & 17544 & Energy & GIFT-Eval \\
\texttt{covid19\_energy} & H & 1 & 1 & 31912 & Energy & GIFT-Eval \\
\texttt{covid\_mobility} & D & 362 & 1 & 148602 & Transport & GIFT-Eval \\
\texttt{dominick} & W & 100014 & 1 & 29652492 & Sales & Chronos \\
\texttt{elecdemand} & 30T & 1 & 1 & 17520 & Energy & GIFT-Eval \\
\texttt{elf} & H & 1 & 1 & 21792 & Energy & GIFT-Eval \\
\texttt{exchange\_rate} & D & 8 & 1 & 84976 & Finance & Chronos \\
\texttt{extended\_web\_traffic\_with\_missing} & D & 145063 & 1 & 370926091 & Web & GIFT-Eval \\
\texttt{godaddy} & M & 3135 & 2 & 128535 & Finance & GIFT-Eval \\
\texttt{hog} & H & 24 & 1 & 421056 & Energy & GIFT-Eval \\
\texttt{ideal} & H & 217 & 1 & 1255253 & Energy & GIFT-Eval \\
\texttt{kaggle\_web\_traffic\_weekly} & W & 145063 & 1 & 16537182 & Web & GIFT-Eval \\
\texttt{lcl} & H & 713 & 1 & 9543553 & Energy & GIFT-Eval \\
\texttt{london\_smart\_meters\_with\_missing} & 30T & 5520 & 1 & 166238880 & Energy & GIFT-Eval \\
\texttt{mexico\_city\_bikes} & H & 494 & 1 & 38687004 & Transport & Chronos \\
\texttt{oikolab\_weather} & H & 8 & 1 & 800456 & Nature & GIFT-Eval \\
\texttt{pdb} & H & 1 & 1 & 17520 & Energy & GIFT-Eval \\
\texttt{pedestrian\_counts} & H & 66 & 1 & 3130762 & Transport & GIFT-Eval \\
\texttt{project\_tycho} & W & 1258 & 1 & 1377707 & Healthcare & GIFT-Eval \\
\texttt{rideshare\_with\_missing} & H & 2304 & 1 & 859392 & Transport & GIFT-Eval \\
\texttt{sceaux} & H & 1 & 1 & 34223 & Energy & GIFT-Eval \\
\texttt{smart} & H & 5 & 1 & 95709 & Energy & GIFT-Eval \\
\texttt{solar\_power} & 4S & 1 & 1 & 7397222 & Energy & GIFT-Eval \\
\texttt{spain} & H & 1 & 1 & 35064 & Energy & GIFT-Eval \\
\texttt{subseasonal} & D & 862 & 4 & 14197140 & Nature & GIFT-Eval \\
\texttt{subseasonal\_precip} & D & 862 & 1 & 9760426 & Nature & GIFT-Eval \\
\texttt{sunspot\_with\_missing} & D & 1 & 1 & 73894 & Nature & GIFT-Eval \\
\texttt{ushcn\_daily} & D & 1218 & 5 & 47080115 & Nature & Chronos \\
\texttt{vehicle\_trips\_with\_missing} & D & 329 & 1 & 32512 & Transport & GIFT-Eval \\
\texttt{weather} & D & 3010 & 1 & 42941700 & Nature & GIFT-Eval \\
\texttt{wiki-rolling\_nips} & D & 47675 & 1 & 40619100 & Web & GIFT-Eval \\
\texttt{wiki\_daily\_100k} & D & 100000 & 1 & 274100000 & Web & Chronos \\
\texttt{wind\_power} & 4S & 1 & 1 & 7397147 & Energy & GIFT-Eval \\
\bottomrule
\end{tabular}
}
\vspace{7pt}
\begin{minipage}{0.88\textwidth}
\footnotesize
\emph{Note.} Frequency aliases follow common time-series conventions:
S = second, T = minute, H = hourly, D = daily, W = weekly,
M = monthly, Q = quarterly, and A = annual.
\end{minipage}
\end{threeparttable}
\end{table}
\renewcommand{\arraystretch}{1}

\section{Dataset Statistics}

This section summarizes the datasets used in our experiments. 
Specifically, Appendix \ref{app:pre-training corpus} describes the corpus used for model pre-training, while Appendix \ref{app:gift-eval} and Appendix \ref{app:fev-eval} present the benchmarks used for downstream evaluation.

\subsection{Pre-training Corpus}
\label{app:pre-training corpus}
Our pre-training corpus comprises both \textbf{real-world} and \textbf{synthetic} time series datasets, covering a broad range of domains and data-generation characteristics.

\textbf{Real-world datasets.}
We aggregate several large-scale time series collections, including the \textsc{GIFT-Eval}~\citep{aksu2024gift} pre-training dataset\footnote{\url{https://huggingface.co/datasets/Salesforce/GiftEvalPretrain}}, the \textsc{Chronos}~\citep{ansari2024chronos} training corpus\footnote{\url{  https://huggingface.co/datasets/autogluon/chronos_datasets}}, and the \textsc{QuitoBench}~\citep{xue2026quitobench} training dataset\footnote{\url{    https://huggingface.co/datasets/hq-bench/quito-corpus}}, as detailed in Table \ref{tab:univariate_pretrain_dataset}. Collectively, these resources span seven major domains: \textit{nature}, \textit{energy}, \textit{transport}, \textit{finance}, \textit{healthcare}, \textit{web}, and \textit{sales}. The resulting corpus contains a large number of \textit{univariate} time series datasets, as well as a small set of \textit{multivariate} time series datasets.

\textbf{Synthetic univariate datasets.}
We also incorporate the synthetic \textit{univariate} datasets introduced by \textsc{Chronos}~\citep{ansari2024chronos}: \textsc{TSMixup} and \textsc{KernelSynth}. \textsc{TSMixup} synthesizes new time series by taking random convex combinations of samples drawn from different real-world datasets, thereby increasing diversity while preserving realistic temporal characteristics. \textsc{KernelSynth}, in contrast, generates synthetic series by randomly composing Gaussian Process (GP) kernels and sampling from the resulting GP priors, producing time series with diverse trends, periodicities, and stochastic patterns.

\textbf{Synthetic multivariate datasets.}
High-quality \textit{multivariate} time series datasets remain relatively scarce in existing public resources. To address this limitation, we construct a large amount of synthetic \textit{multivariate} data through two complementary strategies:
\begin{enumerate}
    \item \textbf{Similarity-based multivariate construction from real univariate series.}
    % Building on the idea of \textsc{TSMixup}, we compute similarities among real univariate time series and group related series to form multivariate datasets.
    Drawing from the real univariate time series presented in Table \ref{tab:univariate_pretrain_dataset}, we compute pairwise similarities to group related sequences into cohesive multivariate datasets. This process enables us to derive multivariate structures from naturally occurring signals while preserving semantic coherence across dimensions.
    
    \item \textbf{Dependency injection over synthetic univariate generators.}
    Inspired by Chronos-2~\citep{ansari2025chronos2}, we transform multiple independently sampled univariate series from base generators (e.g., \textsc{KernelSynth}) into multivariate synthetic time series by imposing explicit dependency structures. These multivariatization procedures include:
    \begin{itemize}[leftmargin=*]
        \item \textbf{Cotemporaneous multivariatizers}, which introduce instantaneous cross-variate dependencies through linear or nonlinear transformations at the same time step;
        \item \textbf{Sequential multivariatizers}, which impose temporal cross-series relations across time, such as \textit{lead--lag} dependencies and \textit{cointegration}.
    \end{itemize}
\end{enumerate}

Through this combination of real-world corpora, \textit{univariate} synthetic generators, and large-scale \textit{multivariate} synthesis, our dataset collection supports training and evaluation across diverse domains and temporal dependency structures.

\subsection{GIFT-Eval Benchmark}
\label{app:gift-eval}
GIFT-Eval is constructed from 15 univariate and 8 multivariate datasets, spanning 7 domains and 10 frequencies. In total, the benchmark contains 144,000 time series and 177 million observations. To support evaluation across forecasting horizons, prediction lengths are determined in two ways. For widely used benchmarks such as M4~\citep{Makridakis2018TheMC}, established prediction lengths are retained. 
For the remaining datasets, the short-horizon prediction length is set to 48 time steps, and the medium- and long-horizon settings are defined according to dataset frequency and domain as 10$\times$ and 15$\times$ the short-horizon length, respectively.
This results in 97 unique combinations of dataset, frequency, and prediction length, with model performance reported as the geometric mean across these configurations.

\renewcommand{\arraystretch}{0.95}
\newcommand{\pw}[2]{#1\,/\,#2}
\begin{table*}[!t]
\caption{Statistics of the GIFT-Eval benchmark across seven domains. Entries under Short-term/Med-term/Long-term are reported as Pred/Win, denoting the prediction length and the number of rolling windows, respectively.}
\label{tab:gift-eval}
\centering
\footnotesize
\setlength{\tabcolsep}{5pt}
\resizebox{\textwidth}{!}{%
\begin{tabular}{clcrrcccc}
\toprule
\textbf{Domain} & \textbf{Dataset Name} & \textbf{Freq.} & \textbf{\#Series} & \textbf{Avg. length} & \textbf{\#Vars} & \textbf{Short-term} & \textbf{Med-term} & \textbf{Long-term} \\
\midrule

\multirow{9}{*}{\rotatebox[origin=c]{90}{\textbf{Nature}}}
& Jena Weather & 10T & 1 & 52,704 & 21 & \pw{48}{20} & \pw{480}{11} & \pw{720}{8} \\
&  & H & 1 & 8,784 & 21 & \pw{48}{19} & \pw{480}{2} & \pw{720}{2} \\
&  & D & 1 & 366 & 21 & \pw{30}{2} & -- & -- \\

& Saugeen & D & 1 & 23,741 & 1 & \pw{30}{20} & -- & -- \\
&  & W-THU & 1 & 3,391 & 1 & \pw{8}{20} & -- & -- \\
&  & M & 1 & 780 & 1 & \pw{12}{7} & -- & -- \\
& Temperature Rain & D & 32,072 & 725 & 1 & \pw{30}{3} & -- & -- \\
& KDD Cup 2018 & H & 270 & 10,898 & 1 & \pw{48}{20} & \pw{480}{2} & \pw{720}{2} \\
&  & D & 270 & 455 & 1 & \pw{30}{2} & -- & -- \\

\cmidrule(lr){1-9}

\multirow{8}{*}{\rotatebox[origin=c]{90}{\textbf{Web/CloudOps}}}
& BizITObs - Application & 10S & 1 & 8,834 & 2 & \pw{60}{15} & \pw{600}{2} & \pw{900}{1} \\
& BizITObs - Service & 10S & 21 & 8,835 & 2 & \pw{60}{15} & \pw{600}{2} & \pw{900}{1} \\
& BizITObs - L2C & 5T & 1 & 31,968 & 7 & \pw{48}{20} & \pw{480}{7} & \pw{720}{5} \\
&  & H & 1 & 2,664 & 7 & \pw{48}{6} & \pw{480}{1} & \pw{720}{1} \\
& Bitbrains - Fast Storage & 5T & 1,250 & 8,640 & 2 & \pw{48}{18} & \pw{480}{2} & \pw{720}{2} \\
&  & H & 1,250 & 721 & 2 & \pw{48}{2} & -- & -- \\
& Bitbrains - rnd & 5T & 500 & 8,640 & 2 & \pw{48}{18} & \pw{480}{2} & \pw{720}{2} \\
&  & H & 500 & 720 & 2 & \pw{48}{2} & -- & -- \\

\cmidrule(lr){1-9}

\multirow{16}{*}{\rotatebox[origin=c]{90}{\textbf{Energy}}}
& ETT1 & 15T & 1 & 69,680 & 7 & \pw{48}{20} & \pw{480}{15} & \pw{720}{10} \\
&  & H & 1 & 17,420 & 7 & \pw{48}{20} & \pw{480}{4} & \pw{720}{3} \\
&  & D & 1 & 725 & 7 & \pw{30}{3} & -- & -- \\
&  & W-THU & 1 & 103 & 7 & \pw{8}{2} & -- & -- \\
& ETT2 & 15T & 1 & 69,680 & 7 & \pw{48}{20} & \pw{480}{15} & \pw{720}{10} \\
&  & H & 1 & 17,420 & 7 & \pw{48}{20} & \pw{480}{4} & \pw{720}{3} \\
&  & D & 1 & 725 & 7 & \pw{30}{3} & -- & -- \\
&  & W-THU & 1 & 103 & 7 & \pw{8}{2} & -- & -- \\
& Solar & 10T & 137 & 52,560 & 1 & \pw{48}{20} & \pw{480}{11} & \pw{720}{8} \\
&  & H & 137 & 8,760 & 1 & \pw{48}{19} & \pw{480}{2} & \pw{720}{2} \\
&  & D & 137 & 365 & 1 & \pw{30}{2} & -- & -- \\
&  & W-FRI & 137 & 52 & 1 & \pw{8}{1} & -- & -- \\
& Electricity & 15T & 370 & 140,256 & 1 & \pw{48}{20} & \pw{480}{20} & \pw{720}{20} \\
&  & H & 370 & 35,064 & 1 & \pw{48}{20} & \pw{480}{8} & \pw{720}{5} \\
&  & D & 370 & 1,461 & 1 & \pw{30}{5} & -- & -- \\
&  & W-FRI & 370 & 208 & 1 & \pw{8}{3} & -- & -- \\

\cmidrule(lr){1-9}

\multirow{7}{*}{\rotatebox[origin=c]{90}{\textbf{Transport}}}
& Loop Seattle & 5T & 323 & 105,120 & 1 & \pw{48}{20} & \pw{480}{20} & \pw{720}{15} \\
&  & H & 323 & 8,760 & 1 & \pw{48}{19} & \pw{480}{2} & \pw{720}{2} \\
&  & D & 323 & 365 & 1 & \pw{30}{2} & -- & -- \\
& SZ-Taxi & 15T & 156 & 2,976 & 1 & \pw{48}{7} & \pw{480}{1} & \pw{720}{1} \\
&  & H & 156 & 744 & 1 & \pw{48}{2} & -- & -- \\
& M\_DENSE & H & 30 & 17,520 & 1 & \pw{48}{20} & \pw{480}{4} & \pw{720}{3} \\
&  & D & 30 & 730 & 1 & \pw{30}{3} & -- & -- \\

\cmidrule(lr){1-9}

\multirow{4}{*}{\rotatebox[origin=c]{90}{\textbf{Sales}}}
& Restaurant & D & 807 & 358 & 1 & \pw{30}{1} & -- & -- \\
& Hierarchical Sales & D & 118 & 1,825 & 1 & \pw{30}{7} & -- & -- \\
&  & W-WED & 118 & 260 & 1 & \pw{8}{4} & -- & -- \\
& Car Parts & M & 2,674 & 51 & 1 & \pw{12}{1} & -- & -- \\

\cmidrule(lr){1-9}

\multirow{6}{*}{\rotatebox[origin=c]{90}{\textbf{Econ/Fin}}}
& M4 Yearly & A & 22,974 & 37 & 1 & \pw{6}{1} & -- & -- \\
& M4 Quarterly & Q & 24,000 & 100 & 1 & \pw{8}{1} & -- & -- \\
& M4 Monthly & M & 48,000 & 234 & 1 & \pw{18}{1} & -- & -- \\
& M4 Weekly & W & 359 & 1,035 & 1 & \pw{13}{1} & -- & -- \\
& M4 Daily & D & 4,227 & 2,371 & 1 & \pw{14}{1} & -- & -- \\
& M4 Hourly & H & 414 & 902 & 1 & \pw{48}{2} & -- & -- \\

\cmidrule(lr){1-9}

\multirow{5}{*}{\rotatebox[origin=c]{90}{\textbf{Healthcare}}}
& Hospital & M & 767 & 84 & 1 & \pw{12}{1} & -- & -- \\
& COVID Deaths & D & 266 & 212 & 1 & \pw{30}{1} & -- & -- \\
& US Births & D & 1 & 7,305 & 1 & \pw{30}{20} & -- & -- \\
&  & W-TUE & 1 & 1,043 & 1 & \pw{8}{14} & -- & -- \\
&  & M & 1 & 240 & 1 & \pw{12}{2} & -- & -- \\

\bottomrule
\end{tabular}%
}

\end{table*}

The benchmark is curated from 10 publicly available sources covering a diverse set of application domains. Below, the included datasets are grouped by domain and described together with their original sources.

\begin{itemize}
\item \textbf{Nature.}
The benchmark includes the Jena Weather dataset\footnote{\url{https://www.bgc-jena.mpg.de/wetter/}}, following the preprocessing protocol used in \textbf{Autoformer}~\citep{Wu2021AutoformerDT}.

\item \textbf{Web/CloudOps.}
This domain contains the BizITObs Application, Service, and L2C datasets\footnote{\url{https://github.com/BizITObs/BizITObservabilityData/tree/main}}, processed according to the pipeline introduced in \textbf{AutoMixer}~\citep{automixer}. These datasets combine business KPIs with IT event channels, forming multivariate time series for observability-related forecasting tasks. In addition, Bitbrains datasets from the \textbf{Grid Workloads Archive}~\citep{grid_workloads_archive} are included in the same domain.

\item \textbf{Sales.}
For the sales domain, the Restaurant dataset is adopted from the \textbf{Recruit Restaurant Forecasting Competition}~\citep{howard2017recruit}, where the objective is to predict future customer visits using reservation and visitation records. Another sales dataset is included from \cite{Mancuso2020AML}.

\item \textbf{Energy.}
The energy domain includes ETT1 and ETT2 from \textbf{Informer}~\citep{zhou2021informer}, which represent electricity transformer temperature and are widely used in long-horizon forecasting. It also includes the Electricity dataset from the \textbf{UCI ML Archive}~\citep{electricityloaddiagrams20112014_321}, containing electricity consumption records for 370 clients, and the Solar dataset from \textbf{LSTNet}~\citep{Lai2017ModelingLA}, which focuses on forecasting solar plant power output.

\item \textbf{Transport.}
Transport datasets are drawn from \textbf{LibCity}~\citep{wang2023libcity}, a benchmark collection of urban spatio-temporal and time series datasets.

\item \textbf{Econ/Fin \& Healthcare.}
A subset of datasets is selected from the \textbf{Monash} repository~\citep{godahewa2021monash}, which provides a broad collection of time series from multiple domains. The selected datasets are chosen to avoid any leakage between pretraining and test data.
\end{itemize}

Detailed dataset statistics are provided in Table \ref{tab:gift-eval}, including frequency, prediction length, variate setting, number of series, series length, and total number of observations. For each time series, the final 10\% of observations is reserved as the test split.

\subsection{fev-bench Benchmark}
\label{app:fev-eval}
The fev-bench benchmark comprises a total of 100 time series forecasting tasks. Detailed dataset statistics are provided in Table \ref{tab:fev-bench}.
This section summarizes the main characteristics of these tasks and provides citations for the corresponding data sources. For datasets originating from forecasting competitions, the benchmark adopts the fixed forecast horizon $T$ specified by the original competition setup. 
For all other datasets, the forecast horizon is determined according to a frequency--horizon mapping. 
An exception is made for a subset of hourly datasets, for which $T=168$ is used in order to support long-range forecasting over a one-week period. 
The number of evaluation windows $W$ is then selected so as to split each series as evenly as possible while ensuring that sufficient historical context remains available for every forecast of length $H$. 
Dataset frequencies are reported using \texttt{pandas} frequency aliases, namely minu\textbf{T}ely, \textbf{H}ourly, \textbf{D}aily, \textbf{W}eekly, \textbf{M}onthly, \textbf{Q}uarterly, and \textbf{Y}early.

The benchmark is constructed from a diverse collection of domains, including macroeconomics, energy systems, retail and sales forecasting, epidemiology, public health, environmental monitoring, and database operations. 
The included datasets can be grouped into the following source categories.

\begin{itemize}

\item \textbf{GIFT-Eval.}
The benchmark includes datasets from the GIFT-Eval corpus \citep{aksu2024gift}, which contains a mixture of univariate and multivariate forecasting tasks. 
The original GIFT-Eval collection draws on data sources compiled from prior benchmark and application papers~\citep{godahewa2021monash,jiang2023libcity,Mancuso2020AML,Wu2021AutoformerDT,automixer}.

\item \textbf{Macroeconomic datasets.}
A broad set of macroeconomic and socioeconomic datasets is included, such as GVAR \citep{mohaddes2024gvar}, US Consumption \citep{WILMS20161256}, Australian Tourism \citep{athanasopoulos2009hierarchical}, FRED-MD \citep{McCracken01102016}, FRED-QD \citep{mccracken2020fred}, world CO$_2$ emissions \citep{Kaggle2025CO2EmissionsByCountry}, life expectancy \citep{Kaggle2025GlobalLifeExpectancy1950_2023}, and global tourism \citep{Kaggle2025TourismEconomicImpact}. For both FRED-MD and FRED-QD, two separate forecasting tasks are defined. The first task follows the CEE model \citep{christiano1999monetary} and focuses on forecasting employment, inflation, and federal funds rate indicators. The second task considers the joint forecasting of 51 core macroeconomic indicators. It should be noted that the benchmark uses the August 2025 snapshot of FRED-MD, which differs from the snapshot used in Monash repository~\citep{godahewa2021monash}.

\item \textbf{Energy datasets.}
The energy-related portion of the benchmark includes several forecasting settings of practical relevance. 
These datasets cover the electricity price forecasting (EPF) benchmark \citep{lago2021forecasting}, ERCOT generation data \citep{ansari2024chronos}, ENTSO-e load data \citep{OPSD2020} paired with weather variates obtained from \texttt{Renewables.ninja} \citep{staffell2023global}, and solar generation data \citep{Kaggle2025RenewableEnergyWeather}. 
Together, these datasets provide a mix of load, price, and renewable generation forecasting tasks.

\item \textbf{BOOMLET.}
The benchmark also includes multivariate observability datasets from BOOMLET \citep{cohen2025toto}, which is itself a subset of the larger BOOM benchmark curated by the original authors. 
To maintain diversity across data sources and prevent overre presentation from a single benchmark family, only BOOMLET datasets with a sampling frequency of at least one minute are retained.

\item \textbf{Forecasting competitions.}
A substantial portion of the benchmark is drawn from forecasting competitions, many of which were hosted on \texttt{kaggle.com}. 
These include Favorita store sales and transactions \citep{Kaggle2020StoreSales}, the M5 competition \citep{makridakis2022m5}, restaurant visitor and reservation forecasting \citep{Kaggle2017RecruitRestaurant}, Rossmann store sales \citep{Kaggle2015Rossmann}, Walmart sales forecasting \citep{Kaggle2014Walmart}, and Rohlik sales forecasting \citep{RohlikSalesForecasting2024}. 
In addition, the benchmark includes the KDD Cup 2022 dataset for wind power forecasting \citep{zhou2022sdwpf}, as well as datasets from the Global Energy Forecasting Competitions held in 2012, 2014, and 2017 \citep{hong2014global}. 
These competition datasets typically come with standardized train--test setups and fixed forecast horizons, making them especially useful for controlled model comparison.

\item \textbf{Other sources.}
To further broaden domain coverage, the benchmark incorporates datasets from several additional sources:
    \begin{itemize}
    \item Influenza-like illness case counts collected by the European Centre for Disease Prevention and Control~\citep{ECDC2025RespiratoryViruses}.
    
    \item Fashion trend data from Hermes~\citep{david2022hermes}.
    
    \item Hospital admissions data from Riyadh~\citep{Kaggle2025RiyadhHospitalAdmissions}.
    
    \item Query count data for Amazon Redshift database servers~\citep{renen2024redset}.
    
    \item Solar energy generation data with associated weather covariates~\citep{Kaggle2025RenewableEnergyWeather}.
    
    \item Air quality measurements from an Italian city together with weather variates~\citep{DeVito2008ElectronicNoseCalibration}.
    
    \item COVID-19 cases, hospital admissions, and deaths in the United Kingdom across multiple administrative levels~\citep{Kaggle2022UKCovidDashboard}.
    \end{itemize}

These additional datasets complement the benchmark by introducing forecasting tasks from healthcare, epidemiology, fashion, environmental sensing, and cloud/database system monitoring, thereby increasing the breadth of real-world scenarios represented in fev-bench.

\end{itemize}

{
\footnotesize
\setlength{\tabcolsep}{8pt}
\renewcommand{\arraystretch}{0.7}

\begin{longtable}{p{4cm}lllccrrr}
\caption{Individual statistics of the fev-bench benchmark across all datasets.}
\label{tab:fev-bench} \\
\toprule
\textbf{Task} & \textbf{Domain} & \textbf{Freq.} & $T$ & $W$ & \textbf{Median length} & \textbf{\# series} & \textbf{\# targets} \\
\midrule
\endfirsthead

\caption[]{Individual statistics of the fev-bench benchmark across all datasets. \textit{(continued)}} \\
\toprule
\textbf{Task} & \textbf{Domain} & \textbf{Freq.} & $H$ & $W$ & \textbf{Median length} & \textbf{\# series} & \textbf{\# targets} \\
\midrule
\endhead

\midrule
\multicolumn{8}{r}{\textit{Continued on next page}} \\
\endfoot

\bottomrule
\endlastfoot

\multicolumn{8}{l}{\textbf{GIFT-Eval}} \\
\midrule
BizITObs-L2C & cloud & 5T & 288 & 20 & 31,968 & 1 & 7 \\
BizITObs-L2C & cloud & H & 24 & 20 & 2,664 & 1 & 7 \\
ETT & energy & 15T & 96 & 20 & 69,680 & 2 & 7 \\
ETT & energy & H & 168 & 20 & 17,420 & 2 & 7 \\
ETT & energy & D & 28 & 20 & 724 & 2 & 7 \\
ETT & energy & W & 13 & 5 & 103 & 2 & 7 \\
Hierarchical Sales & retail & D & 28 & 10 & 1,825 & 118 & 1 \\
Hierarchical Sales & retail & W & 13 & 10 & 260 & 118 & 1 \\
Hospital & healthcare & M & 12 & 4 & 84 & 767 & 1 \\
Jena Weather & nature & 10T & 144 & 20 & 52,704 & 1 & 21 \\
Jena Weather & nature & D & 28 & 11 & 366 & 1 & 21 \\
Jena Weather & nature & H & 24 & 20 & 8,784 & 1 & 21 \\
Loop Seattle & mobility & D & 28 & 10 & 365 & 323 & 1 \\
Loop Seattle & mobility & 5T & 288 & 10 & 105,120 & 323 & 1 \\
Loop Seattle & mobility & H & 168 & 10 & 8,760 & 323 & 1 \\
M-DENSE & mobility & D & 28 & 10 & 730 & 30 & 1 \\
M-DENSE & mobility & H & 168 & 10 & 17,520 & 30 & 1 \\
SZ Taxi & mobility & 15T & 96 & 10 & 2,976 & 156 & 1 \\
SZ Taxi & mobility & H & 168 & 2 & 744 & 156 & 1 \\
Solar & energy & W & 13 & 1 & 52 & 137 & 1 \\
Solar & energy & D & 28 & 10 & 365 & 137 & 1 \\

\midrule
\multicolumn{8}{l}{\textbf{Macroeconomic datasets}} \\
\midrule
Australian Tourism & econ & Q & 8 & 2 & 36 & 89 & 1 \\
FRED-MD-CEE & econ & M & 12 & 20 & 798 & 1 & 3 \\
FRED-MD-Macro & econ & M & 12 & 20 & 798 & 1 & 51 \\
FRED-QD-CEE & econ & Q & 8 & 20 & 266 & 1 & 3 \\
FRED-QD-Macro & econ & Q & 8 & 20 & 266 & 1 & 51 \\
GVAR & econ & Q & 8 & 10 & 178 & 33 & 6 \\
US Consumption & econ & M & 12 & 10 & 792 & 31 & 1 \\
US Consumption & econ & Q & 8 & 10 & 262 & 31 & 1 \\
US Consumption & econ & Y & 5 & 10 & 64 & 31 & 1 \\
World CO2 Emissions & econ & Y & 5 & 9 & 60 & 191 & 1 \\
World Life Expectancy & econ & Y & 5 & 10 & 74 & 237 & 1 \\
World Tourism & econ & Y & 5 & 2 & 21 & 178 & 1 \\

\midrule
\multicolumn{8}{l}{\textbf{Energy datasets}} \\
\midrule
ENTSO-e Load & energy & 15T & 96 & 20 & 175,292 & 6 & 1 \\
ENTSO-e Load & energy & 30T & 96 & 20 & 87,645 & 6 & 1 \\
ENTSO-e Load & energy & H & 168 & 20 & 43,822 & 6 & 1 \\
EPF-BE & energy & H & 24 & 20 & 52,416 & 1 & 1 \\
EPF-DE & energy & H & 24 & 20 & 52,416 & 1 & 1 \\
EPF-FR & energy & H & 24 & 20 & 52,416 & 1 & 1 \\
EPF-NP & energy & H & 24 & 20 & 52,416 & 1 & 1 \\
EPF-PJM & energy & H & 24 & 20 & 52,416 & 1 & 1 \\
ERCOT & energy & D & 28 & 20 & 6,452 & 8 & 1 \\
ERCOT & energy & H & 168 & 20 & 154,872 & 8 & 1 \\
ERCOT & energy & M & 12 & 15 & 211 & 8 & 1 \\
ERCOT & energy & W & 13 & 20 & 921 & 8 & 1 \\
GFC12 & energy & H & 168 & 10 & 39,414 & 11 & 1 \\
GFC14 & energy & H & 168 & 20 & 17,520 & 1 & 1 \\
GFC17 & energy & H & 168 & 20 & 17,544 & 8 & 1 \\
Solar with Weather & energy & 15T & 96 & 20 & 198,600 & 1 & 1 \\
Solar with Weather & energy & H & 24 & 20 & 49,648 & 1 & 1 \\

\midrule
\multicolumn{8}{l}{\textbf{BOOMLET}} \\
\midrule
BOOMLET-1062 & cloud & 5T & 288 & 20 & 16,384 & 1 & 21 \\
BOOMLET-1209 & cloud & 5T & 288 & 20 & 16,384 & 1 & 53 \\
BOOMLET-1225 & cloud & T & 60 & 20 & 16,384 & 1 & 49 \\
BOOMLET-1230 & cloud & 5T & 288 & 20 & 16,384 & 1 & 23 \\
BOOMLET-1282 & cloud & T & 60 & 20 & 16,384 & 1 & 35 \\
BOOMLET-1487 & cloud & 5T & 288 & 20 & 16,384 & 1 & 54 \\
BOOMLET-1631 & cloud & 30T & 96 & 20 & 10,463 & 1 & 40 \\
BOOMLET-1676 & cloud & 30T & 96 & 20 & 10,463 & 1 & 100 \\
BOOMLET-1855 & cloud & H & 24 & 20 & 5,231 & 1 & 52 \\
BOOMLET-1975 & cloud & H & 24 & 20 & 5,231 & 1 & 75 \\
BOOMLET-2187 & cloud & H & 24 & 20 & 5,231 & 1 & 100 \\
BOOMLET-285 & cloud & T & 60 & 20 & 16,384 & 1 & 75 \\
BOOMLET-619 & cloud & T & 60 & 20 & 16,384 & 1 & 52 \\
BOOMLET-772 & cloud & T & 60 & 20 & 16,384 & 1 & 67 \\
BOOMLET-963 & cloud & T & 60 & 20 & 16,384 & 1 & 28 \\

\midrule
\multicolumn{8}{l}{\textbf{Forecasting competitions}} \\
\midrule
Favorita Store Sales & retail & M & 12 & 2 & 54 & 1,579 & 1 \\
Favorita Store Sales & retail & W & 13 & 10 & 240 & 1,579 & 1 \\
Favorita Store Sales & retail & D & 28 & 10 & 1,688 & 1,579 & 1 \\
Favorita Transactions & retail & M & 12 & 2 & 54 & 51 & 1 \\
Favorita Transactions & retail & W & 13 & 10 & 240 & 51 & 1 \\
Favorita Transactions & retail & D & 28 & 10 & 1,688 & 51 & 1 \\
KDD Cup 2022 & energy & D & 14 & 10 & 243 & 134 & 1 \\
KDD Cup 2022 & energy & 10T & 288 & 10 & 35,279 & 134 & 1 \\
KDD Cup 2022 & energy & 30T & 96 & 10 & 11,758 & 134 & 1 \\
M5 & retail & M & 12 & 1 & 58 & 30,490 & 1 \\
M5 & retail & W & 13 & 1 & 257 & 30,490 & 1 \\
M5 & retail & D & 28 & 1 & 1,810 & 30,490 & 1 \\
Restaurant & retail & D & 28 & 8 & 296 & 817 & 1 \\
Rohlik Orders & retail & W & 8 & 5 & 170 & 7 & 1 \\
Rohlik Orders & retail & D & 61 & 5 & 1,197 & 7 & 1 \\
Rohlik Sales & retail & W & 8 & 1 & 150 & 5,243 & 1 \\
Rohlik Sales & retail & D & 14 & 1 & 1,046 & 5,390 & 1 \\
Rossmann & retail & W & 13 & 8 & 133 & 1,115 & 1 \\
Rossmann & retail & D & 48 & 10 & 942 & 1,115 & 1 \\
Walmart & retail & W & 39 & 1 & 143 & 2,936 & 1 \\

\midrule
\multicolumn{8}{l}{\textbf{Other datasets}} \\
\midrule
ECDC ILI & healthcare & W & 13 & 10 & 201 & 25 & 1 \\
Hermes & retail & W & 52 & 1 & 261 & 10,000 & 1 \\
Hospital Admissions & healthcare & D & 28 & 20 & 1,731 & 8 & 1 \\
Hospital Admissions & healthcare & W & 13 & 16 & 246 & 8 & 1 \\
Redset & cloud & 5T & 288 & 10 & 25,920 & 118 & 1 \\
Redset & cloud & 15T & 96 & 10 & 8,640 & 126 & 1 \\
Redset & cloud & H & 24 & 10 & 2,160 & 138 & 1 \\
UCI Air Quality & nature & H & 168 & 20 & 9,357 & 1 & 4 \\
UCI Air Quality & nature & D & 28 & 11 & 389 & 1 & 4 \\
UK COVID-Nation-Cumulative & healthcare & D & 28 & 20 & 729 & 4 & 3 \\
UK COVID-Nation-Cumulative & healthcare & W & 8 & 4 & 105 & 4 & 3 \\
UK COVID-Nation-New & healthcare & D & 28 & 20 & 729 & 4 & 3 \\
UK COVID-Nation-New & healthcare & W & 8 & 4 & 105 & 4 & 3 \\
UK COVID-UTLA-Cumulative & healthcare & W & 13 & 5 & 104 & 214 & 1 \\
UK COVID-UTLA-New & healthcare & D & 28 & 10 & 721 & 214 & 1 \\

\end{longtable}
}

\section{Sampling Details}
\label{app:sampling details}

\subsection{Flexible Context Length and Horizon Sampling}
Unlike decoder-only transformers that inherently support variable context and prediction lengths during pre-training, encoder-only architectures face significant challenges in achieving such flexible length settings, which is quite crucial for model generalization. Theoretically, while models pre-trained with fixed target horizon can perform arbitrary-length inference via auto-regression, they are susceptible to the cumulative error propagation common in decoder-only structures. To address this, we implement flexible context length and horizon during the sampling stage. Specifically, for each sampled entity $e_i$ with context length $L_i$ and target horizon $T_i$, we construct each batch by left-padding input contexts with \texttt{NaN} values to the batch-wise maximum input length $L_{\text{max}} = {\max} \{L_i\}_{i=1}^N$. Meanwhile, target outputs are right-padded to a pre-defined maximum horizon $T_{\text{max}} = 480$, corresponding to 30 output tokens with patch length $L_p=16$. The padded context positions, together with invalid observations, constitute the binary observation mask $\mathcal{M}$, while the padded target segments are excluded from the final loss calculation. Such flexible sampling strategy effectively enhances the predictive generalization of \NAME.

\subsection{Runtime Multivariate Sampling}
\label{app:runtime multivariate sampling}
To mitigate potential GPU memory bottlenecks arising from multivariate sampling, we adopt a dynamic variate sampling strategy at runtime. After flexible context length and horizon sampling, for each entity $e_i$ with $v_i$ variates, we first randomly permute the variate order to reduce order bias and encourage permutation-robust, content-driven cross-variate modeling, consistent with prior findings that channel shuffling improves robustness to channel ordering in multivariate forecasting~\citep{xu2026cpiri}. Then we traverse variates in the permuted order. Subsequently, we iteratively append valid variates to the training sample until all candidates are processed or a pre-defined per-sample limit, $M_{\max}$, is reached. 
Finally, entity $e_i$ contributes $m_i = \min(v_i, M_{\max})$ variates. This runtime design naturally supports heterogeneous variate dimensionalities and prevents dimensionality-related computational bottlenecks while facing excessively large $v_i$.

After variate selection, multivariate samples may still vary in variate count. To batch them efficiently, we enforce a batch-level variate budget and accumulate samples until the total number of retained variates reaches a preset threshold, stabilizing memory usage across training steps. 
% Context and label sequences are NaN-padded, with the corresponding input and loss masks zero-padded accordingly. Rather than packing data into a dense tensor of shape $(B, V_{\max}, L)$, 
% We concatenate all valid variates across samples and produce a total number of $M$ variates for each batch. We additionally maintain an entity ID for each variate to track its originating sample. 
This variate-wise batching strategy substantially reduces channel-padding waste and enables efficient training on multivariate data with highly variate dimensionality.

\section{Additional Visualization}

\begin{figure}[!htb]
    \centering
    \includegraphics[width=\linewidth]{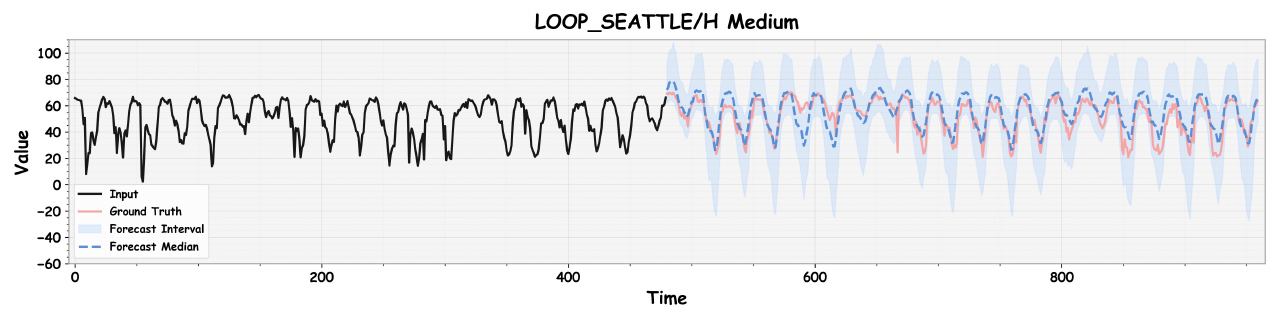}
    \caption{Medium-horizon ($T=480$) quantile forecasts on Loop Seattle.}
    \label{fig:loop_seattle_h}
\end{figure}

\begin{figure}[!htb]
    \centering
    \includegraphics[width=\linewidth]{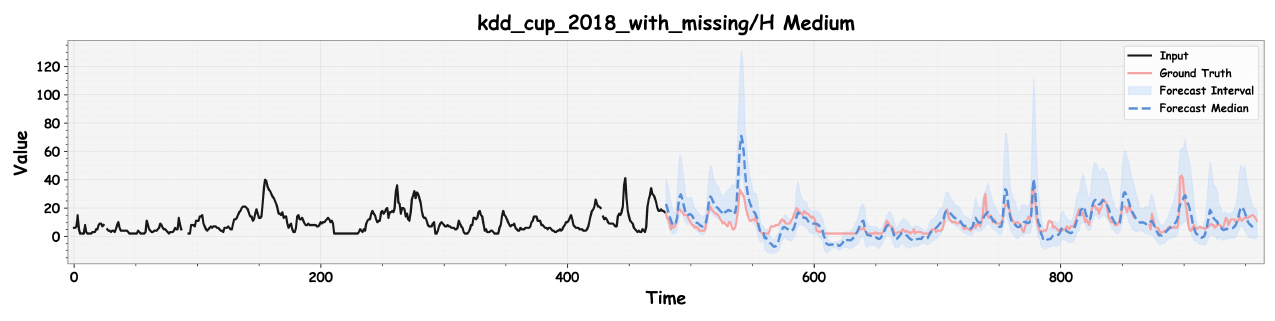}
    \caption{Medium-horizon ($T=480$) quantile forecasts on kdd cup 2018.}
    \label{fig:kdd_cup_H}
\end{figure}

\begin{figure}[!htb]
    \centering
    \includegraphics[width=0.98\linewidth]{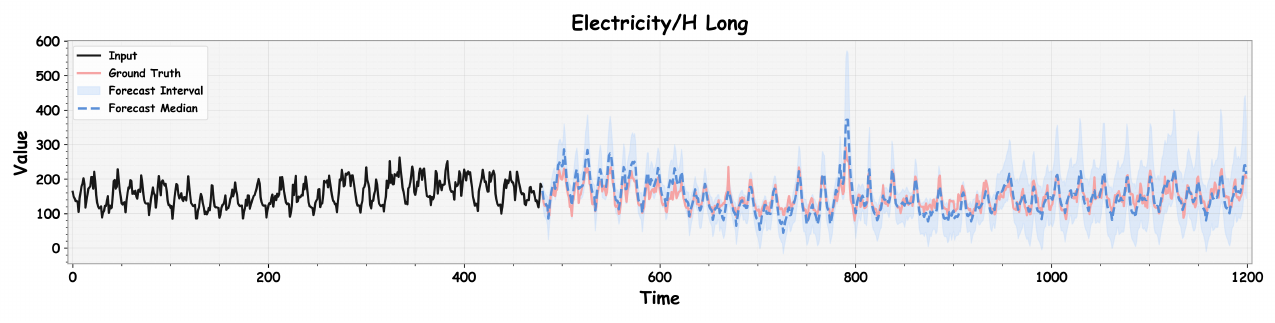}
    \caption{Long-horizon ($T=720$) quantile forecasts on Electricity.}
    \label{fig:ecl_h}
\end{figure}

\begin{figure}[!htb]
    \centering
    \includegraphics[width=0.98\linewidth]{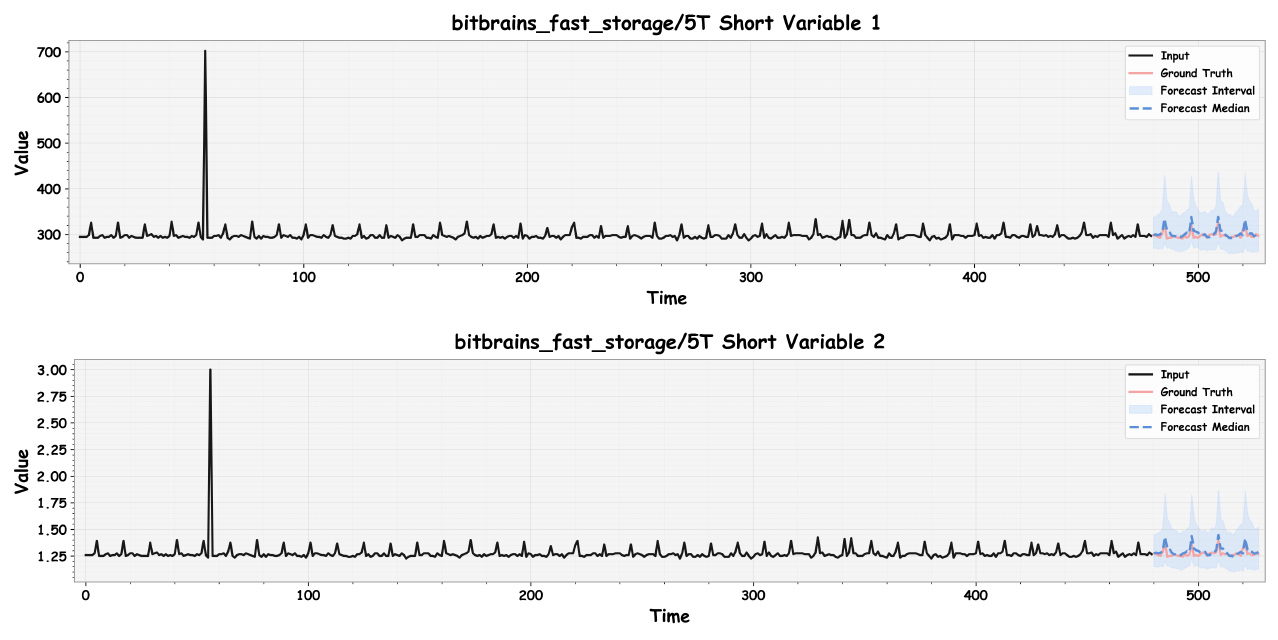}
    \caption{Short-horizon ($T=48$) quantile forecasts on bitbrains fast storage.}
    \label{fig:bit_fast_5t}
\end{figure}

\begin{figure}[!htb]
    \centering
    \includegraphics[width=0.98\linewidth]{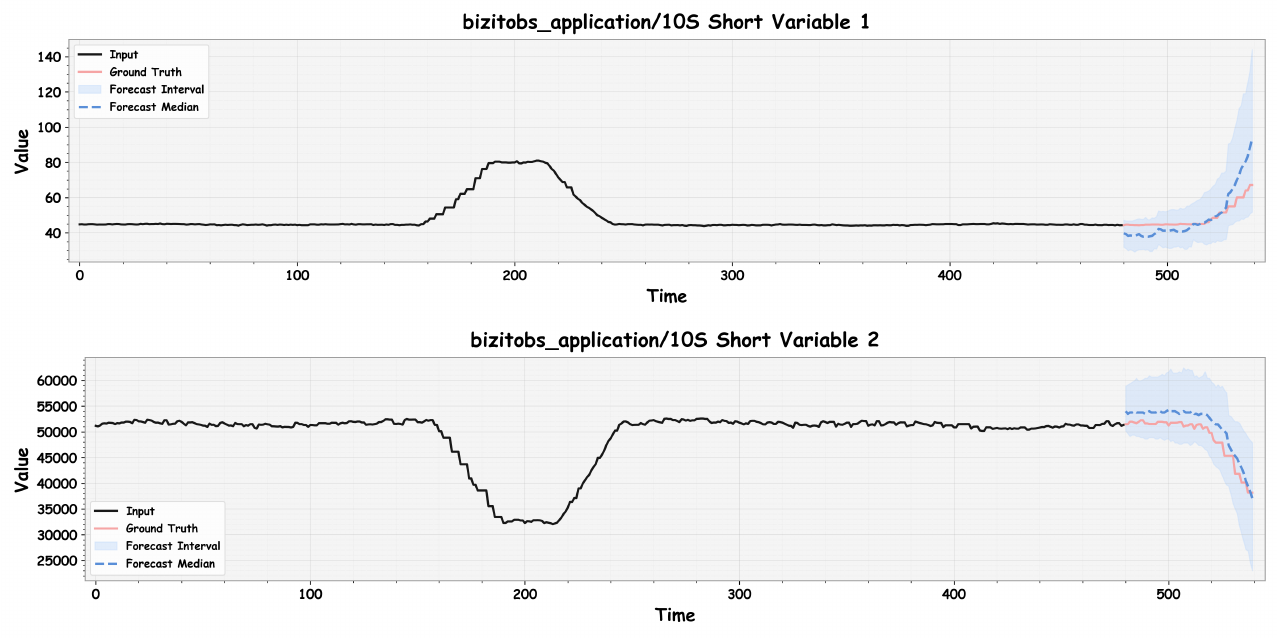}
    \caption{Short-horizon ($T=60$) quantile forecasts on bizitobs application.}
    \label{fig:biz_application_10s}
\end{figure}

We provide visualizations of \NAME's quantile forecasts across representative datasets at different frequencies \{5T, 15T, 10S, H\} and prediction horizons $T=\{48, 60, 480, 720\}$. Specifically, Figure \ref{fig:loop_seattle_h} shows medium-horizon ($T=480$) forecasts for Loop Seattle/H, Figure\ref{fig:kdd_cup_H} for KDD Cup 2018/H, and Figure \ref{fig:ecl_h} for Electricity/H at $T=720$. Figure \ref{fig:bit_fast_5t} presents forecasts on Bitbrains Fast Storage/5T at $T=48$, where the two channels exhibit strong positive correlation, which \NAME accurately captures. In contrast, Figure \ref{fig:biz_application_10s} shows forecasts on Bizitobs Application/10s at $T=60$, where two channels are negatively correlated, and \NAME successfully models the inverse relationship. These results demonstrate that the variable-to-prototype design effectively captures both positive and negative inter-variable dependencies, and preserves temporal consistency across diverse datasets. Moreover, the visualizations highlight \NAME's robustness in handling different sampling frequencies and prediction horizons without manual adjustment.

\begin{figure}[!t]
    \centering
    \includegraphics[width=0.82\linewidth]{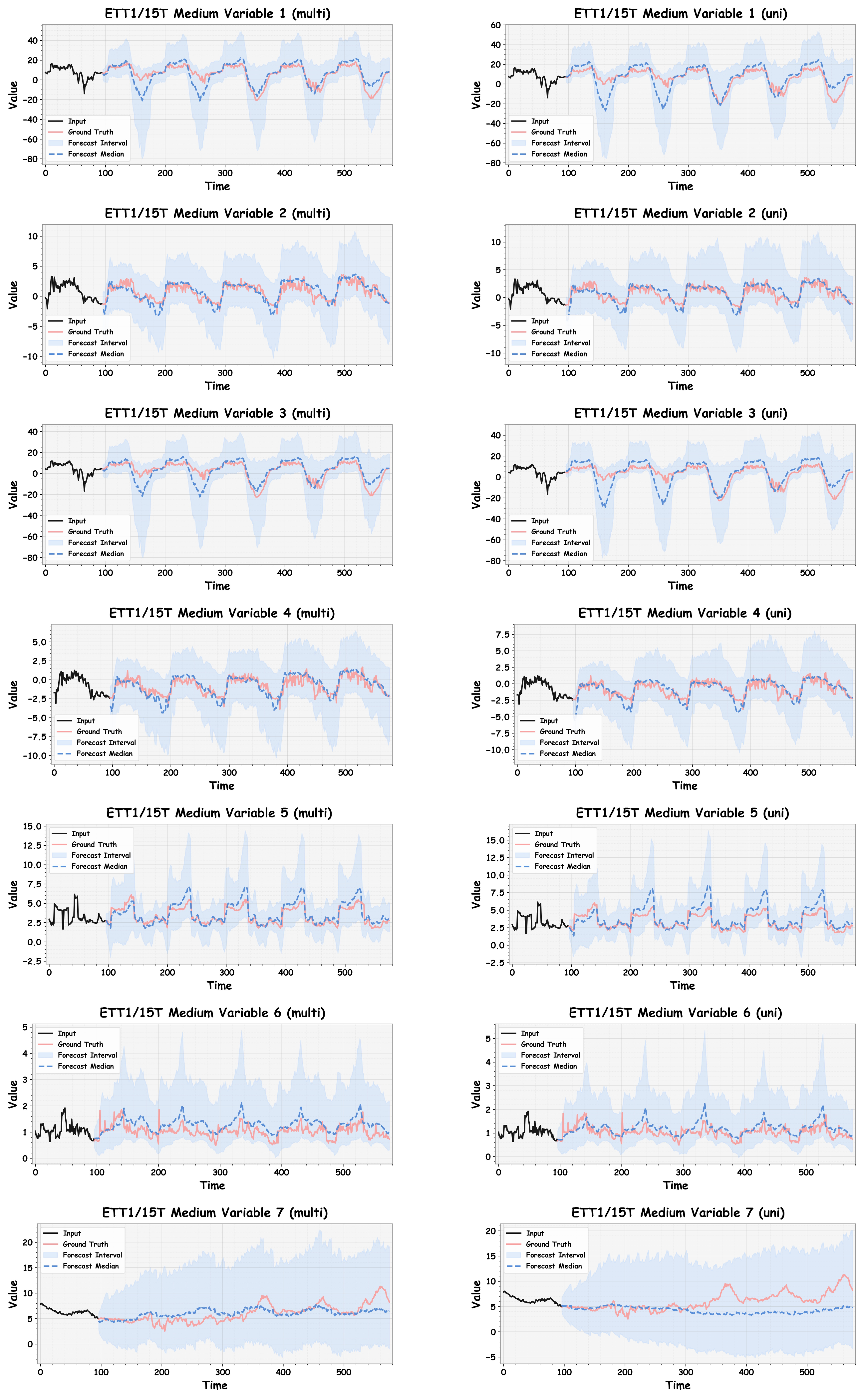}
    \caption{Medium-horizon ($T=480$) quantile forecasts on ETT1/15T with 7 channels, comparing \NAME with and without multivariate inference.}
    \label{fig:ett1_15t_full}
\end{figure}

\begin{figure}[!t]
    \centering
    \includegraphics[width=0.82\linewidth]{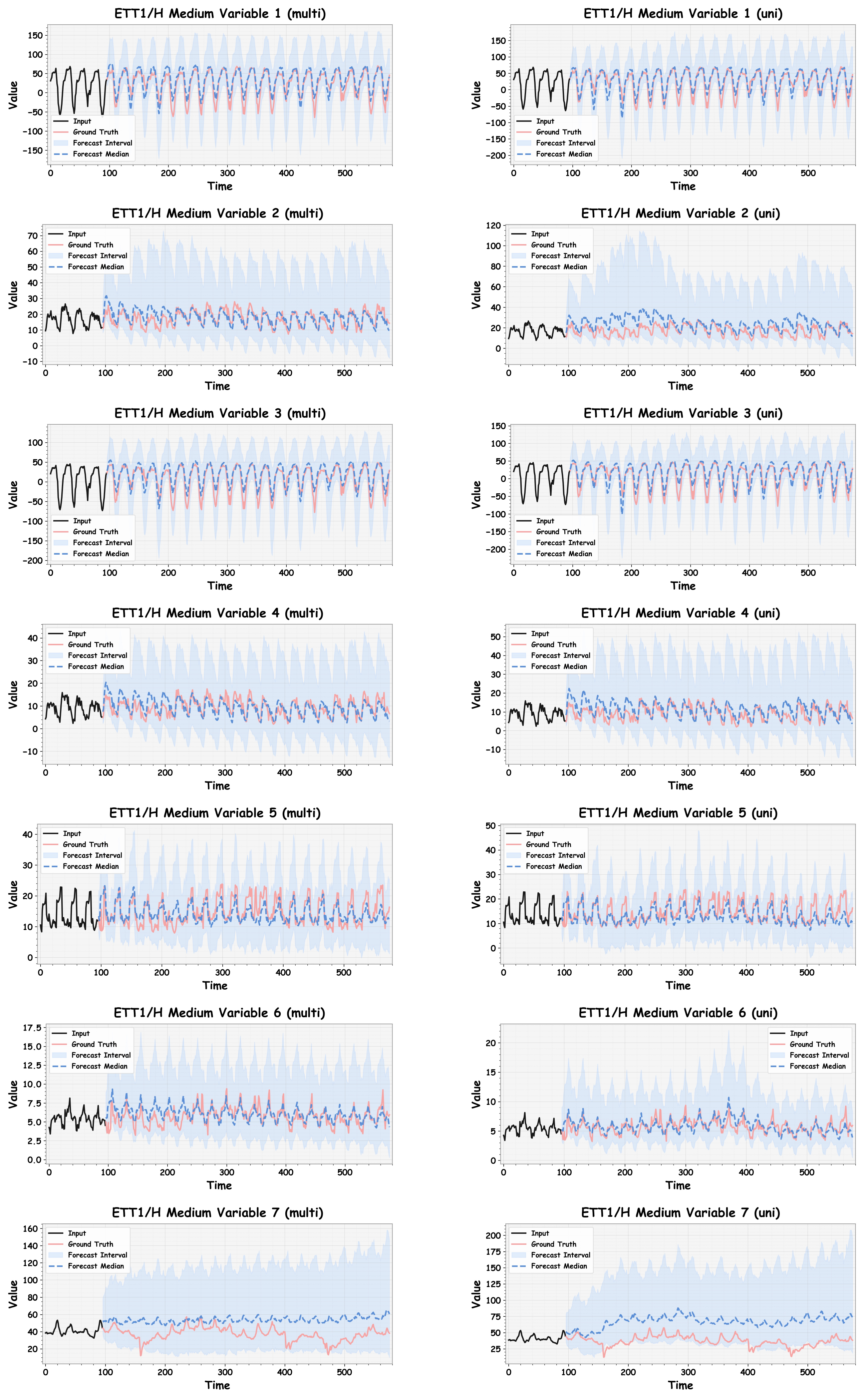}
    \caption{Medium-horizon ($T=480$) quantile forecasts on ETT1/H with 7 channels, comparing \NAME with and without multivariate inference.}
    \label{fig:ett1_h_full}
\end{figure}

Moreover, we visualize \NAME’s medium-horizon ($T=480$) quantile forecasts on ETTh1/15T and ETTh1/H, comparing results with and without multivariate inference across all seven channels. As shown in Figure \ref{fig:ett1_15t_full} and Figure \ref{fig:ett1_h_full}, enabling multivariate inference allows \NAME to more accurately capture complex inter-variable relationships, including both strong and subtle dependencies, which results in tighter predictive intervals and improved alignment with observed dynamics across channels.

%% file: main.bbl
\begin{thebibliography}{65}
\providecommand{\natexlab}[1]{#1}
\providecommand{\url}[1]{\texttt{#1}}
\expandafter\ifx\csname urlstyle\endcsname\relax
  \providecommand{\doi}[1]{doi: #1}\else
  \providecommand{\doi}{doi: \begingroup \urlstyle{rm}\Url}\fi

\bibitem[Admin and Cukierski(2014)]{Kaggle2014Walmart}
Walmart~Competition Admin and Will Cukierski.
\newblock Walmart recruiting - store sales forecasting.
\newblock \url{https://kaggle.com/competitions/walmart-recruiting-store-sales-forecasting}, 2014.
\newblock Kaggle.

\bibitem[Aksu et~al.(2024)Aksu, Woo, Liu, Liu, Liu, Savarese, Xiong, and Sahoo]{aksu2024gift}
Taha Aksu, Gerald Woo, Juncheng Liu, Xu~Liu, Chenghao Liu, Silvio Savarese, Caiming Xiong, and Doyen Sahoo.
\newblock Gift-{EVAL}: A benchmark for general time series forecasting model evaluation.
\newblock \emph{arXiv preprint arXiv:2410.10393}, 2024.

\bibitem[Ansari et~al.(2024)Ansari, Stella, Turkmen, Zhang, Mercado, Shen, Shchur, Rangapuram, Arango, Kapoor, et~al.]{ansari2024chronos}
Abdul~Fatir Ansari, Lorenzo Stella, Caner Turkmen, Xiyuan Zhang, Pedro Mercado, Huibin Shen, Oleksandr Shchur, Syama~Sundar Rangapuram, Sebastian~Pineda Arango, Shubham Kapoor, et~al.
\newblock Chronos: Learning the language of time series.
\newblock \emph{Transactions on Machine Learning Research}, 2024.

\bibitem[Ansari et~al.(2025)Ansari, Shchur, Küken, Auer, Han, Mercado, Rangapuram, Shen, Stella, Zhang, Goswami, Kapoor, Maddix, Guerron, Hu, Yin, Erickson, Desai, Wang, Rangwala, Karypis, Wang, and Bohlke-Schneider]{ansari2025chronos2}
Abdul~Fatir Ansari, Oleksandr Shchur, Jaris Küken, Andreas Auer, Boran Han, Pedro Mercado, Syama~Sundar Rangapuram, Huibin Shen, Lorenzo Stella, Xiyuan Zhang, Mononito Goswami, Shubham Kapoor, Danielle~C. Maddix, Pablo Guerron, Tony Hu, Junming Yin, Nick Erickson, Prateek~Mutalik Desai, Hao Wang, Huzefa Rangwala, George Karypis, Yuyang Wang, and Michael Bohlke-Schneider.
\newblock Chronos-2: From univariate to universal forecasting.
\newblock \emph{arXiv preprint arXiv:2510.15821}, 2025.

\bibitem[Athanasopoulos et~al.(2009)Athanasopoulos, Ahmed, and Hyndman]{athanasopoulos2009hierarchical}
George Athanasopoulos, Roman~A. Ahmed, and Rob~J. Hyndman.
\newblock Hierarchical forecasts for {A}ustralian domestic tourism.
\newblock \emph{International Journal of Forecasting}, 25\penalty0 (1):\penalty0 146--166, January 2009.
\newblock ISSN 0169-2070.
\newblock \doi{10.1016/j.ijforecast.2008.07.004}.
\newblock \url{http://dx.doi.org/10.1016/j.ijforecast.2008.07.004}.

\bibitem[Auer et~al.(2025)Auer, Podest, Klotz, B{\"o}ck, Klambauer, and Hochreiter]{auertirex}
Andreas Auer, Patrick Podest, Daniel Klotz, Sebastian B{\"o}ck, G{\"u}nter Klambauer, and Sepp Hochreiter.
\newblock Tirex: Zero-shot forecasting across long and short horizons with enhanced in-context learning.
\newblock In \emph{Neural Information Processing Systems}, 2025.

\bibitem[Ba et~al.(2016)Ba, Kiros, and Hinton]{layernorm}
Jimmy~Lei Ba, Jamie~Ryan Kiros, and Geoffrey~E Hinton.
\newblock Layer normalization.
\newblock In \emph{arXiv preprint arXiv:1607.06450}, 2016.

\bibitem[Christiano et~al.(1999)Christiano, Eichenbaum, and Evans]{christiano1999monetary}
Lawrence~J. Christiano, Martin Eichenbaum, and Charles~L. Evans.
\newblock Monetary policy shocks: What have we learned and to what end?
\newblock In \emph{Handbook of Macroeconomics}, volume~1 of \emph{Handbook of Macroeconomics}, pages 65--148. Elsevier, 1999.
\newblock \doi{https://doi.org/10.1016/S1574-0048(99)01005-8}.
\newblock \url{https://www.sciencedirect.com/science/article/pii/S1574004899010058}.

\bibitem[Cohen et~al.(2025)Cohen, Khwaja, Doubli, Lemaachi, Lettieri, Masson, Miccinilli, Ram{\'e}, Ren, Rostamizadeh, et~al.]{cohen2025toto}
Ben Cohen, Emaad Khwaja, Youssef Doubli, Salahidine Lemaachi, Chris Lettieri, Charles Masson, Hugo Miccinilli, Elise Ram{\'e}, Qiqi Ren, Afshin Rostamizadeh, et~al.
\newblock This time is different: An observability perspective on time series foundation models.
\newblock In \emph{Neural Information Processing Systems}, 2025.

\bibitem[Das et~al.(2024)Das, Kong, Sen, and Zhou]{das2024decoder}
Abhimanyu Das, Weihao Kong, Rajat Sen, and Yichen Zhou.
\newblock A decoder-only foundation model for time-series forecasting.
\newblock In \emph{International Conference on Machine Learning}, pages 10148--10167, 2024.

\bibitem[Data(2020)]{OPSD2020}
Open Power~System Data.
\newblock Data package time series. version 2020-10-06, 2020.
\newblock \url{https://doi.org/10.25832/time_series/2020-10-06}.

\bibitem[data from official {UK}~government sources(2022)]{Kaggle2022UKCovidDashboard}
{UK COVID-19} data from official {UK}~government sources.
\newblock {UK COVID-19} dashboard data.
\newblock \url{https://www.kaggle.com/datasets/happyadam73/uk-covid19-dashboard-data-sqlite-compressed}, 2022.
\newblock Kaggle.

\bibitem[David et~al.(2022)David, Bellot, and Corff]{david2022hermes}
Etienne David, Jean Bellot, and Sylvain~Le Corff.
\newblock H{ERMES}: Hybrid error-corrector model with inclusion of external signals for nonstationary fashion time series.
\newblock \emph{arXiv preprint arXiv:2202.03224}, 2022.

\bibitem[{De Vito} et~al.(2008){De Vito}, Massera, Piga, Martinotto, and {Di Francia}]{DeVito2008ElectronicNoseCalibration}
S.~{De Vito}, E.~Massera, M.~Piga, L.~Martinotto, and G.~{Di Francia}.
\newblock On field calibration of an electronic nose for benzene estimation in an urban pollution monitoring scenario.
\newblock \emph{Sensors and Actuators B: Chemical}, 129\penalty0 (2):\penalty0 750--757, 2008.
\newblock ISSN 0925-4005.
\newblock \doi{https://doi.org/10.1016/j.snb.2007.09.060}.
\newblock \url{https://www.sciencedirect.com/science/article/pii/S0925400507007691}.

\bibitem[{ECDC}(2025)]{ECDC2025RespiratoryViruses}
{ECDC}.
\newblock Respiratory viruses weekly data.
\newblock \url{https://github.com/EU-ECDC/Respiratory_viruses_weekly_data/tree/main}, 2025.
\newblock Open data repository; weekly respiratory virus surveillance in the {EU/EEA}.

\bibitem[Fleming and Wallace(1986)]{lago2021forecasting}
Philip~J Fleming and John~J Wallace.
\newblock How not to lie with statistics: the correct way to summarize benchmark results.
\newblock \emph{Communications of the ACM}, 29\penalty0 (3):\penalty0 218--221, 1986.

\bibitem[FlorianKnauer and Cukierski(2015)]{Kaggle2015Rossmann}
FlorianKnauer and Will Cukierski.
\newblock Rossmann store sales.
\newblock \url{https://kaggle.com/competitions/rossmann-store-sales}, 2015.
\newblock Kaggle.

\bibitem[Godahewa et~al.(2021)Godahewa, Bergmeir, Webb, Hyndman, and Montero-Manso]{godahewa2021monash}
Rakshitha~Wathsadini Godahewa, Christoph Bergmeir, Geoffrey~I. Webb, Rob Hyndman, and Pablo Montero-Manso.
\newblock Monash time series forecasting archive.
\newblock In \emph{The Conference on Neural Information Processing Systems Datasets and Benchmarks Track}, 2021.
\newblock \url{https://openreview.net/forum?id=wEc1mgAjU-}.

\bibitem[Goswami et~al.(2024)Goswami, Szafer, Choudhry, Cai, Li, and Dubrawski]{goswamimoment}
Mononito Goswami, Konrad Szafer, Arjun Choudhry, Yifu Cai, Shuo Li, and Artur Dubrawski.
\newblock M{OMENT}: A family of open time-series foundation models.
\newblock In \emph{International Conference on Machine Learning}, 2024.

\bibitem[Hong et~al.(2014)Hong, Pinson, and Fan]{hong2014global}
Tao Hong, Pierre Pinson, and Shu Fan.
\newblock Global energy forecasting competition 2012.
\newblock \emph{International Journal of Forecasting}, 30\penalty0 (2):\penalty0 357--363, 2014.

\bibitem[Hoo et~al.(2025)Hoo, M{\"u}ller, Salinas, and Hutter]{hoo2025tables}
Shi~Bin Hoo, Samuel M{\"u}ller, David Salinas, and Frank Hutter.
\newblock From tables to time: Extending tabpfn-v2 to time series forecasting.
\newblock \emph{arXiv preprint arXiv:2501.02945}, 2025.

\bibitem[Howard et~al.(2017{\natexlab{a}})Howard, Yui, McDonald, and Cukierski]{Kaggle2017RecruitRestaurant}
Addison Howard, Haruka Yui, Mark McDonald, and Will Cukierski.
\newblock Recruit restaurant visitor forecasting.
\newblock \url{https://www.kaggle.com/c/recruit-restaurant-visitor-forecasting}, 2017{\natexlab{a}}.
\newblock Kaggle.

\bibitem[Howard et~al.(2017{\natexlab{b}})Howard, Yui, McDonald, and Cukierski]{howard2017recruit}
Addison Howard, Haruka Yui, Mark McDonald, and Will Cukierski.
\newblock Recruit restaurant visitor forecasting.
\newblock \url{https://kaggle.com/competitions/recruit-restaurant-visitor-forecasting}, 2017{\natexlab{b}}.
\newblock Kaggle.

\bibitem[Jiang et~al.(2023)Jiang, Han, Jiang, Zhao, and Wang]{jiang2023libcity}
Jiawei Jiang, Chengkai Han, Wenjun Jiang, Wayne~Xin Zhao, and Jingyuan Wang.
\newblock Libcity: A unified library towards efficient and comprehensive urban spatial-temporal prediction.
\newblock \emph{arXiv preprint arXiv:2304.14343}, 2023.

\bibitem[Jin et~al.(2023)Jin, Wang, Ma, Chu, Zhang, Shi, Chen, Liang, Li, Pan, et~al.]{jintime}
Ming Jin, Shiyu Wang, Lintao Ma, Zhixuan Chu, James~Y Zhang, Xiaoming Shi, Pin-Yu Chen, Yuxuan Liang, Yuan-Fang Li, Shirui Pan, et~al.
\newblock Time-{LLM}: Time series forecasting by reprogramming large language models.
\newblock In \emph{International Conference on Learning Representations}, 2023.

\bibitem[Kong et~al.(2025)Kong, Chen, Liu, Ning, Zhang, Muhammad~Marier, Liu, Chen, and Xia]{survey}
Xiangjie Kong, Zhenghao Chen, Weiyao Liu, Kaili Ning, Lechao Zhang, Syauqie Muhammad~Marier, Yichen Liu, Yuhao Chen, and Feng Xia.
\newblock Deep learning for time series forecasting: a survey.
\newblock \emph{International Journal of Machine Learning and Cybernetics}, 16\penalty0 (7):\penalty0 5079--5112, 2025.

\bibitem[Kottapalli et~al.(2025)Kottapalli, Hubli, Chandrashekhara, Jain, Hubli, Botla, and Doddaiah]{foundation_survey}
Siva Rama~Krishna Kottapalli, Karthik Hubli, Sandeep Chandrashekhara, Garima Jain, Sunayana Hubli, Gayathri Botla, and Ramesh Doddaiah.
\newblock Foundation models for time series: A survey.
\newblock \emph{arXiv preprint arXiv:2504.04011}, 2025.

\bibitem[Lai et~al.(2017)Lai, Chang, Yang, and Liu]{Lai2017ModelingLA}
Guokun Lai, Wei-Cheng Chang, Yiming Yang, and Hanxiao Liu.
\newblock Modeling long- and short-term temporal patterns with deep neural networks.
\newblock In \emph{The International ACM SIGIR Conference on Research \& Development in Information Retrieval}, 2017.
\newblock \url{https://api.semanticscholar.org/CorpusID:4922476}.

\bibitem[lexis Cook et~al.(2020)lexis Cook, DanB, inversion, and Holbrook]{Kaggle2020StoreSales}
lexis Cook, DanB, inversion, and Ryan Holbrook.
\newblock Store sales -- time series forecasting.
\newblock \url{https://www.kaggle.com/competitions/store-sales-time-series-forecasting}, 2020.
\newblock Kaggle.

\bibitem[Liu et~al.(2025{\natexlab{a}})Liu, Aksu, Liu, Liu, Yan, Pham, Savarese, Sahoo, Xiong, and Li]{liu2025moirai}
Chenghao Liu, Taha Aksu, Juncheng Liu, Xu~Liu, Hanshu Yan, Quang Pham, Silvio Savarese, Doyen Sahoo, Caiming Xiong, and Junnan Li.
\newblock Moirai 2.0: When less is more for time series forecasting.
\newblock \emph{arXiv preprint arXiv:2511.11698}, 2025{\natexlab{a}}.

\bibitem[Liu et~al.(2024)Liu, Zhang, Li, Huang, Wang, and Long]{liu2024timer}
Yong Liu, Haoran Zhang, Chenyu Li, Xiangdong Huang, Jianmin Wang, and Mingsheng Long.
\newblock Timer: generative pre-trained transformers are large time series models.
\newblock In \emph{International Conference on Machine Learning}, pages 32369--32399, 2024.

\bibitem[Liu et~al.(2025{\natexlab{b}})Liu, Qin, Shi, Chen, Yang, Huang, Wang, and Long]{liu2025sundial}
Yong Liu, Guo Qin, Zhiyuan Shi, Zhi Chen, Caiyin Yang, Xiangdong Huang, Jianmin Wang, and Mingsheng Long.
\newblock Sundial: A family of highly capable time series foundation models.
\newblock In \emph{International Conference on Machine Learning}, pages 39295--39317. PMLR, 2025{\natexlab{b}}.

\bibitem[Liu et~al.(2026)Liu, Su, Wang, Zhang, Liu, Wang, Ye, Xiang, Wang, and Long]{liu2026timers1}
Yong Liu, Xingjian Su, Shiyu Wang, Haoran Zhang, Haixuan Liu, Yuxuan Wang, Zhou Ye, Yang Xiang, Jianmin Wang, and Mingsheng Long.
\newblock Timer-{S}1: A billion-scale time series foundation model with serial scaling.
\newblock \emph{arXiv preprint arXiv:2603.04791}, 2026.

\bibitem[Loshchilov and Hutter(2019)]{loshchilov2018adamw}
Ilya Loshchilov and Frank Hutter.
\newblock Decoupled weight decay regularization.
\newblock In \emph{International Conference on Learning Representations}, 2019.
\newblock \url{https://openreview.net/forum?id=Bkg6RiCqY7}.

\bibitem[Makridakis et~al.(2018)Makridakis, Spiliotis, and Assimakopoulos]{Makridakis2018TheMC}
Spyros Makridakis, Evangelos Spiliotis, and Vassilios Assimakopoulos.
\newblock The {M}4 competition: Results, findings, conclusion and way forward.
\newblock \emph{International Journal of Forecasting}, 2018.

\bibitem[Makridakis et~al.(2022)Makridakis, Spiliotis, and Assimakopoulos]{makridakis2022m5}
Spyros Makridakis, Evangelos Spiliotis, and Vassilios Assimakopoulos.
\newblock M5 accuracy competition: Results, findings, and conclusions.
\newblock \emph{International Journal of Forecasting}, 38\penalty0 (4):\penalty0 1346--1364, 2022.
\newblock ISSN 0169-2070.
\newblock \doi{https://doi.org/10.1016/j.ijforecast.2021.11.013}.
\newblock \url{https://www.sciencedirect.com/science/article/pii/S0169207021001874}.
\newblock Special Issue: M5 competition.

\bibitem[Mancuso et~al.(2021)Mancuso, Piccialli, and Sudoso]{Mancuso2020AML}
Paolo Mancuso, Veronica Piccialli, and Antonio~M Sudoso.
\newblock A machine learning approach for forecasting hierarchical time series.
\newblock \emph{Expert Systems with Applications}, 182:\penalty0 115102, 2021.

\bibitem[Maverick(2025)]{Kaggle2025RenewableEnergyWeather}
AI~Maverick.
\newblock Renewable energy and weather conditions.
\newblock \url{https://www.kaggle.com/datasets/samanemami/renewable-energy-and-weather-conditions}, 2025.
\newblock Kaggle.

\bibitem[McCracken and Ng(2016)]{McCracken01102016}
Michael~W. McCracken and Serena Ng.
\newblock F{RED-MD}: A monthly database for macroeconomic research.
\newblock \emph{Journal of Business \& Economic Statistics}, 34\penalty0 (4):\penalty0 574--589, 2016.
\newblock \doi{10.1080/07350015.2015.1086655}.
\newblock \url{https://doi.org/10.1080/07350015.2015.1086655}.

\bibitem[McCracken and Ng(2021)]{mccracken2020fred}
Michael~W. McCracken and Serena Ng.
\newblock F{RED-QD}: A quarterly database for macroeconomic research.
\newblock \emph{Review}, 103\penalty0 (1):\penalty0 1--44, January 2021.
\newblock \doi{10.20955/r.103.1-44}.
\newblock \url{https://ideas.repec.org/a/fip/fedlrv/90588.html}.

\bibitem[MichalKecera(2024)]{RohlikSalesForecasting2024}
MichalKecera.
\newblock Rohlik sales forecasting challenge.
\newblock \url{https://kaggle.com/competitions/rohlik-sales-forecasting-challenge-v2}, 2024.
\newblock Kaggle.

\bibitem[Mohaddes and Raissi(2024)]{mohaddes2024gvar}
Kamiar Mohaddes and Mehdi Raissi.
\newblock Compilation, revision and updating of the global var (gvar) database.
\newblock Mendeley Data, Version 1, 2024.
\newblock \url{https://doi.org/10.17632/kfp5fhgkvf.1}.

\bibitem[Nie et~al.(2023)Nie, Nguyen, Sinthong, and Kalagnanam]{nietime}
Yuqi Nie, Nam~H Nguyen, Phanwadee Sinthong, and Jayant Kalagnanam.
\newblock A time series is worth 64 words: Long-term forecasting with transformers.
\newblock In \emph{International Conference on Learning Representations}, 2023.

\bibitem[Noor(2025)]{Kaggle2025GlobalLifeExpectancy1950_2023}
Nafay~Un Noor.
\newblock Global life expectancy data (1950--2023).
\newblock \url{https://www.kaggle.com/datasets/nafayunnoor/global-life-expectancy-data-1950-2023}, 2025.
\newblock Kaggle.

\bibitem[of~Health~Affairs and Ministry~of Health(2024)]{Kaggle2025RiyadhHospitalAdmissions}
General~Directorate of~Health~Affairs and Saudi~Arabia Ministry~of Health.
\newblock Riyadh hospital admissions dataset (2020–2024).
\newblock \url{https://www.kaggle.com/dsv/9992619}, 2024.

\bibitem[Palaskar et~al.(2024)Palaskar, Ekambaram, Jati, Gantayat, Saha, Nagar, Nguyen, Dayama, Sindhgatta, Mohapatra, Kumar, Kalagnanam, Hemachandra, and Rangaraj]{automixer}
Santosh Palaskar, Vijay Ekambaram, Arindam Jati, Neelamadhav Gantayat, Avirup Saha, Seema Nagar, Nam Nguyen, Pankaj Dayama, Renuka Sindhgatta, Prateeti Mohapatra, Harshit Kumar, Jayant Kalagnanam, Nandyala Hemachandra, and Narayan Rangaraj.
\newblock Automixer for improved multivariate time-series forecasting on business and it observability data.
\newblock \emph{Proceedings of the AAAI Conference on Artificial Intelligence}, 38:\penalty0 22962--22968, 2024.

\bibitem[Pedersen(2025)]{Kaggle2025CO2EmissionsByCountry}
Ulrik~Thyge Pedersen.
\newblock {CO2} emissions by country.
\newblock \url{https://www.kaggle.com/datasets/ulrikthygepedersen/co2-emissions-by-country}, 2025.
\newblock Kaggle.

\bibitem[Qurban(2025)]{Kaggle2025TourismEconomicImpact}
Bushra Qurban.
\newblock Tourism and economic impact.
\newblock \url{https://www.kaggle.com/datasets/bushraqurban/tourism-and-economic-impact}, 2025.
\newblock Kaggle.

\bibitem[Shchur et~al.(2025)Shchur, Ansari, Turkmen, Stella, Erickson, Guerron, Bohlke-Schneider, and Wang]{shchur2025fev}
Oleksandr Shchur, Abdul~Fatir Ansari, Caner Turkmen, Lorenzo Stella, Nick Erickson, Pablo Guerron, Michael Bohlke-Schneider, and Yuyang Wang.
\newblock fev-bench: A realistic benchmark for time series forecasting.
\newblock \emph{arXiv preprint arXiv:2509.26468}, 2025.

\bibitem[Shen et~al.(2015)Shen, Van~Beek, and Iosup]{grid_workloads_archive}
Siqi Shen, Vincent Van~Beek, and Alexandru Iosup.
\newblock Statistical characterization of business-critical workloads hosted in cloud datacenters.
\newblock In \emph{IEEE/ACM International Symposium on Cluster, Cloud and Grid Computing}, pages 465--474. IEEE, 2015.

\bibitem[Shoeybi et~al.(2019)Shoeybi, Patwary, Puri, LeGresley, Casper, and Catanzaro]{megatron-lm}
Mohammad Shoeybi, Mostofa Patwary, Raul Puri, Patrick LeGresley, Jared Casper, and Bryan Catanzaro.
\newblock Megatron-{LM}: Training multi-billion parameter language models using model parallelism.
\newblock \emph{arXiv preprint arXiv:1909.08053}, 2019.

\bibitem[Staffell et~al.(2023)Staffell, Pfenninger, and Johnson]{staffell2023global}
Iain Staffell, Stefan Pfenninger, and Nathan Johnson.
\newblock A global model of hourly space heating and cooling demand at multiple spatial scales.
\newblock \emph{Nature Energy}, 8\penalty0 (12):\penalty0 1328--1344, 2023.
\newblock \doi{10.1038/s41560-023-01341-5}.
\newblock \url{https://doi.org/10.1038/s41560-023-01341-5}.

\bibitem[Trindade(2015)]{electricityloaddiagrams20112014_321}
Artur Trindade.
\newblock {ElectricityLoadDiagrams20112014}.
\newblock UCI Machine Learning Repository, 2015.
\newblock {DOI}: https://doi.org/10.24432/C58C86.

\bibitem[van Renen et~al.(2024)van Renen, Horn, Pfeil, Vaidya, Dong, Narayanaswamy, Liu, Saxena, Kipf, and Kraska]{renen2024redset}
Alexander van Renen, Dominik Horn, Pascal Pfeil, Kapil Vaidya, Wenjian Dong, Murali Narayanaswamy, Zhengchun Liu, Gaurav Saxena, Andreas Kipf, and Tim Kraska.
\newblock Why {TPC} is not enough: An analysis of the amazon redshift fleet.
\newblock \emph{Proc. VLDB Endow.}, 17\penalty0 (11):\penalty0 3694–3706, July 2024.
\newblock ISSN 2150-8097.
\newblock \doi{10.14778/3681954.3682031}.
\newblock \url{https://doi.org/10.14778/3681954.3682031}.

\bibitem[Vaswani et~al.(2017)Vaswani, Shazeer, Parmar, Uszkoreit, Jones, Gomez, Kaiser, and Polosukhin]{vaswani2017attention}
Ashish Vaswani, Noam Shazeer, Niki Parmar, Jakob Uszkoreit, Llion Jones, Aidan~N Gomez, {\L}ukasz Kaiser, and Illia Polosukhin.
\newblock Attention is all you need.
\newblock \emph{Advances in neural information processing systems}, 30, 2017.

\bibitem[Wang et~al.(2023)Wang, Jiang, Jiang, Han, and Zhao]{wang2023libcity}
Jingyuan Wang, Jiawei Jiang, Wenjun Jiang, Chengkai Han, and Wayne~Xin Zhao.
\newblock Towards efficient and comprehensive urban spatial-temporal prediction: A unified library and performance benchmark.
\newblock \emph{arXiv preprint arXiv:2304.14343}, 2023.

\bibitem[Wilms and Croux(2016)]{WILMS20161256}
Ines Wilms and Christophe Croux.
\newblock Forecasting using sparse cointegration.
\newblock \emph{International Journal of Forecasting}, 32\penalty0 (4):\penalty0 1256--1267, 2016.
\newblock ISSN 0169-2070.
\newblock \doi{https://doi.org/10.1016/j.ijforecast.2016.04.005}.
\newblock \url{https://www.sciencedirect.com/science/article/pii/S0169207016300589}.

\bibitem[Woo et~al.(2024)Woo, Liu, Kumar, Xiong, Savarese, and Sahoo]{woo2024unified}
Gerald Woo, Chenghao Liu, Akshat Kumar, Caiming Xiong, Silvio Savarese, and Doyen Sahoo.
\newblock Unified training of universal time series forecasting transformers.
\newblock In \emph{International Conference on Machine Learning}, 2024.

\bibitem[Wu et~al.(2021)Wu, Xu, Wang, and Long]{Wu2021AutoformerDT}
Haixu Wu, Jiehui Xu, Jianmin Wang, and Mingsheng Long.
\newblock Autoformer: Decomposition transformers with auto-correlation for long-term series forecasting.
\newblock In \emph{Neural Information Processing Systems}, 2021.
\newblock \url{https://api.semanticscholar.org/CorpusID:235623791}.

\bibitem[Xiaoming et~al.(2025)Xiaoming, Shiyu, Yuqi, Dianqi, Zhou, Qingsong, and Jin]{xiaoming2025time}
Shi Xiaoming, Wang Shiyu, Nie Yuqi, Li~Dianqi, Ye~Zhou, Wen Qingsong, and Ming Jin.
\newblock Time-{M}o{E}: Billion-scale time series foundation models with mixture of experts.
\newblock In \emph{International Conference on Learning Representations}, 2025.

\bibitem[Xu et~al.(2026)Xu, Zhang, Jing, Nie, Chen, and Zhang]{xu2026cpiri}
Jiyuan Xu, Wenyu Zhang, Xin Jing, Jiahao Nie, Shuai Chen, and Shuai Zhang.
\newblock {CP}i{R}i: Channel permutation-invariant relational interaction for multivariate time series forecasting.
\newblock In \emph{The International Conference on Learning Representations}, 2026.
\newblock \url{https://openreview.net/forum?id=tgnXCCjKE3}.

\bibitem[Xue et~al.(2026)Xue, Zhu, Zhang, Cai, Wang, Mu, Zhou, Li, Di, and Yu]{xue2026quitobench}
Siqiao Xue, Zhaoyang Zhu, Wei Zhang, Rongyao Cai, Rui Wang, Yixiang Mu, Fan Zhou, Jianguo Li, Peng Di, and Hang Yu.
\newblock Quito{B}ench: A high-quality open time series forecasting benchmark.
\newblock \emph{arXiv preprint arXiv:2603.26017}, 2026.

\bibitem[Ye et~al.(2025)Ye, Dong, Xia, Sun, Zhu, Huang, and Wei]{diffattn}
Tianzhu Ye, Li~Dong, Yuqing Xia, Yutao Sun, Yi~Zhu, Gao Huang, and Furu Wei.
\newblock Differential transformer.
\newblock In \emph{International Conference on Learning Representations}, 2025.

\bibitem[Zhou et~al.(2021)Zhou, Zhang, Peng, Zhang, Li, Xiong, and Zhang]{zhou2021informer}
Haoyi Zhou, Shanghang Zhang, Jieqi Peng, Shuai Zhang, Jianxin Li, Hui Xiong, and Wancai Zhang.
\newblock Informer: Beyond efficient transformer for long sequence time-series forecasting.
\newblock In \emph{Proceedings of the AAAI conference on artificial intelligence}, volume~35, pages 11106--11115, 2021.

\bibitem[Zhou et~al.(2024)Zhou, Lu, Xiao, Tang, Su, Li, Liu, Lyu, Ma, and Dou]{zhou2022sdwpf}
Jingbo Zhou, Xinjiang Lu, Yixiong Xiao, Jian Tang, Jiantao Su, Yu~Li, Ji~Liu, Junfu Lyu, Yanjun Ma, and Dejing Dou.
\newblock S{DWPF}: A dataset for spatial dynamic wind power forecasting over a large turbine array.
\newblock \emph{Scientific Data}, 11\penalty0 (1):\penalty0 649, 2024.
\newblock \doi{10.1038/s41597-024-03427-5}.
\newblock \url{https://doi.org/10.1038/s41597-024-03427-5}.

\end{thebibliography}
